\pdfoutput=1
\documentclass{article}
\usepackage[table]{xcolor}
\definecolor{elsrow}{RGB}{235, 243, 252}

\usepackage{enumitem}

\usepackage[preprint]{neurips_2026}

\usepackage{microtype}
\usepackage{graphicx}
\usepackage{subcaption}

\usepackage{booktabs}
\usepackage{array}
\usepackage{multirow}

\usepackage[most]{tcolorbox}
\usepackage{fvextra}

\usepackage{amsmath}
\usepackage{amssymb}

\usepackage{placeins}
\usepackage{caption}

\usepackage{hyperref}
\hypersetup{colorlinks=true, linkcolor=blue!60!black, citecolor=blue!60!black, urlcolor=magenta!60!black,
  pdftitle={Extraneous Cognitive Load in Large Language Models}, pdfauthor={Arushi Gupta, Sheng Liu, James Zou}}
\usepackage[capitalize,noabbrev]{cleveref}

\newtcolorbox{promptbox}[2][]{%
  enhanced, breakable,
  colback=gray!3, colframe=black!15,
  boxrule=0.6pt, arc=6pt,
  left=10pt,right=10pt,top=8pt,bottom=8pt,
  title=\textbf{#2}, fonttitle=\small, coltitle=black,
  attach boxed title to top left={xshift=10pt,yshift=-2mm},
  boxed title style={
    colback=white,colframe=black!15,arc=6pt,boxrule=0.6pt,
    left=6pt,right=6pt,top=2pt,bottom=2pt
  },
  #1
}

\definecolor{lightBlue}{rgb}{0.78, 0.85, 1.0}
\definecolor{lightRed}{rgb}{1.0, 0.85, 0.85}

\newtcbox{\bluebox}{on line, box align=base, colback=lightBlue,colframe=white,
  size=fbox,arc=3pt, before upper=\strut, top=-2pt, bottom=-4pt,
  left=-2pt, right=-2pt, boxrule=0pt}
\newtcbox{\redbox}{on line, box align=base, colback=lightRed,colframe=white,
  size=fbox,arc=3pt, before upper=\strut, top=-2pt, bottom=-4pt,
  left=-2pt, right=-2pt, boxrule=0pt}

\DeclareRobustCommand{\dashifted}{\raisebox{0.1ex}{\tiny$\downarrow$}}
\DeclareRobustCommand{\upshifted}{\raisebox{0.1ex}{\tiny$\uparrow$}}
\newcommand{\dar}[1]{{\scriptsize\redbox{\dashifted{#1}}}}
\newcommand{\uab}[1]{{\scriptsize\bluebox{\upshifted{#1}}}}

\title{Extraneous Cognitive Load in Large Language Models}

\author{%
  Arushi Gupta \\ Stanford University \\ \texttt{arugupta@stanford.edu} \And
  Sheng Liu \\ Stanford University \\ \texttt{shengl@stanford.edu} \And
  James Zou \\ Stanford University \\ \texttt{jamesz@stanford.edu}
}

\begin{document}

\maketitle
{\renewcommand{\thefootnote}{}\footnotetext{Code and data: \url{https://github.com/arushig100/CogLoadBench}}}

\begin{abstract}

Cognitive Load Theory (CLT) shows that human problem-solving is affected by both intrinsic load, arising from task difficulty, and extraneous load, arising from how task information is presented. Large language models (LLMs) show analogous effects: presentation alone can affect model performance, but these effects have not yet been systematically isolated from task difficulty. To address this gap, we introduce CogLoadBench, a controlled benchmark that varies five CLT-inspired presentation factors across three reasoning task families. Across models, we find that higher extraneous load conditions often reduce accuracy by either adding competition or complicating the reasoning path. To support diagnosis and mitigation of load-related effects, we propose the Extraneous Load Score (ELS), a prompt-time metric computed from activations that estimates a model's extraneous load, and show that it generalizes across reasoning task types and load levels. We further show that shifting model representations toward lower-load directions at inference time can improve average model performance without changing prompt content.
\end{abstract}

\section{Introduction}
\label{sec:introduction}
Human performance on reasoning tasks depends on both the complexity of the task itself and how the task is presented. Cognitive Load Theory (CLT), a framework from cognitive psychology, formalizes this relationship by distinguishing two key types of human cognitive load when problem-solving:
\begin{itemize}
    \item \textbf{Intrinsic load}, which comes from the difficulty of the task itself \\
    \textit{Example}: adding two numbers vs. solving a long multi-step math problem
    \item \textbf{Extraneous load}, which comes from how the task is presented \\
    \textit{Example}: a clearly stated math question vs. the same question surrounded by irrelevant text
\end{itemize}

Human performance consistently declines when either intrinsic or extraneous load increases. These effects are well established across domains such as learning, memory, and decision making \citep{clt2}. Similar patterns have also been observed in large language models (LLMs) when they are prompted at inference time. When a prompt requires more complex reasoning (higher intrinsic load) or inefficiently presents information (higher extraneous load), performance generally decreases.

Most prior work on LLMs has focused on their ability to handle increasingly complex reasoning, which is analogous to intrinsic load. In contrast, much less attention has been given to analogous extraneous load factors. Understanding how different forms of extraneous load affect LLM behavior is important for three reasons. \textbf{First, it helps us explain model failures more accurately}: a model may fail because the prompt adds unnecessary load, not because the task itself is too difficult. \textbf{Second, it supports targeted diagnosis and mitigation}: different kinds of extraneous load can affect models in different ways, and understanding a model’s specific weaknesses allows us to design prompt or model changes that address them. \textbf{Third, it is important for assessing deployment reliability}: real prompts are often long, messy, or poorly organized, so models must be robust to extraneous load-like conditions to perform well in practice.

\begin{figure}[tb]
  \centering
  \includegraphics[width=\linewidth]{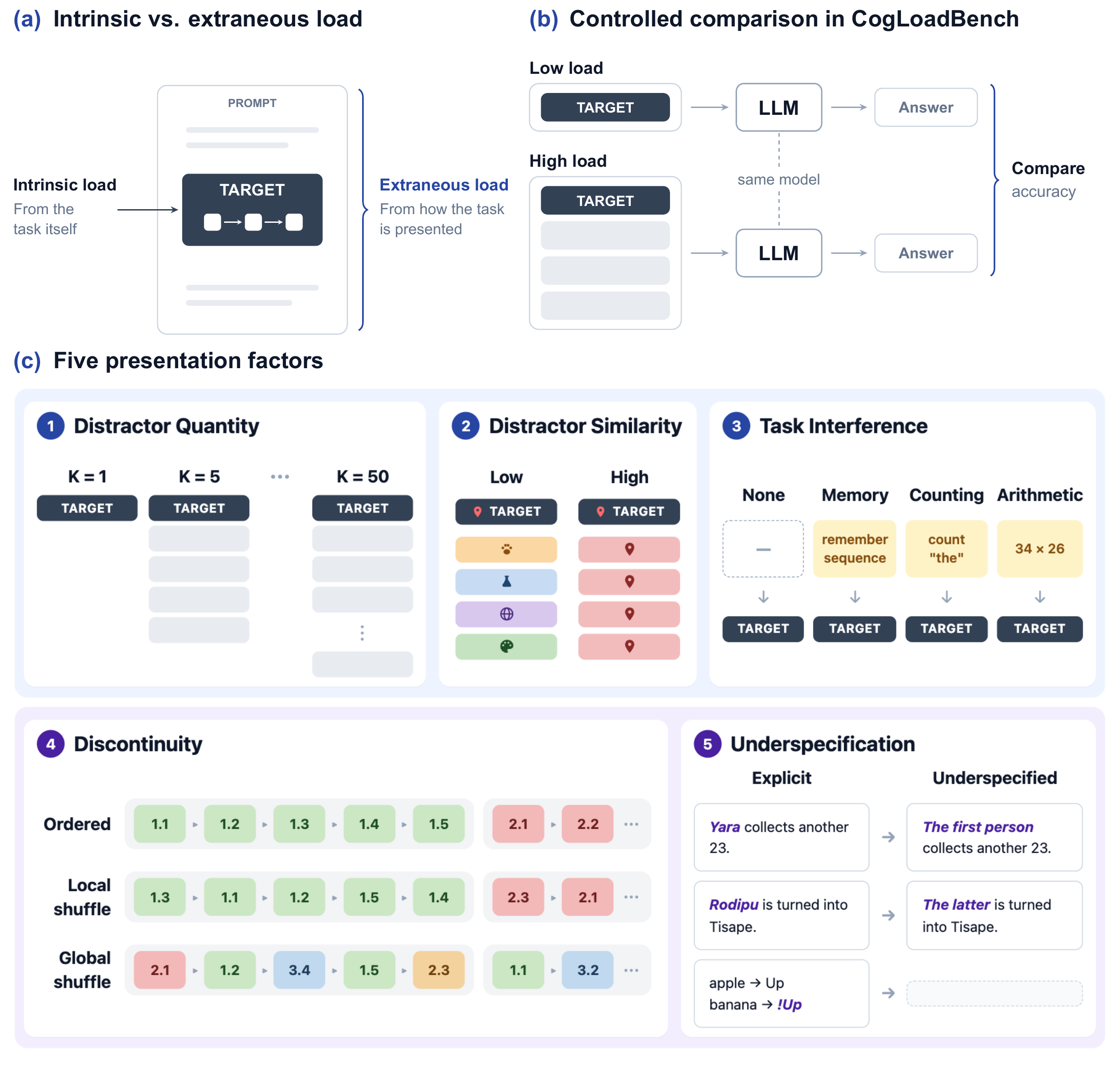}
  \caption{\textbf{Isolating the effects of prompt presentation with CogLoadBench.} \textbf{(a)} Intrinsic load comes from the task itself; extraneous load comes from how it is presented. \textbf{(b)} CogLoadBench keeps the target problem fixed and changes only its presentation, so comparing a model's accuracy under low and high load isolates the effect of extraneous load. \textbf{(c)} We vary five presentation factors, one at a time.}
  \label{fig:overview}
\end{figure}

\begin{figure}[tb]
  \centering
  \includegraphics[width=0.90\linewidth]{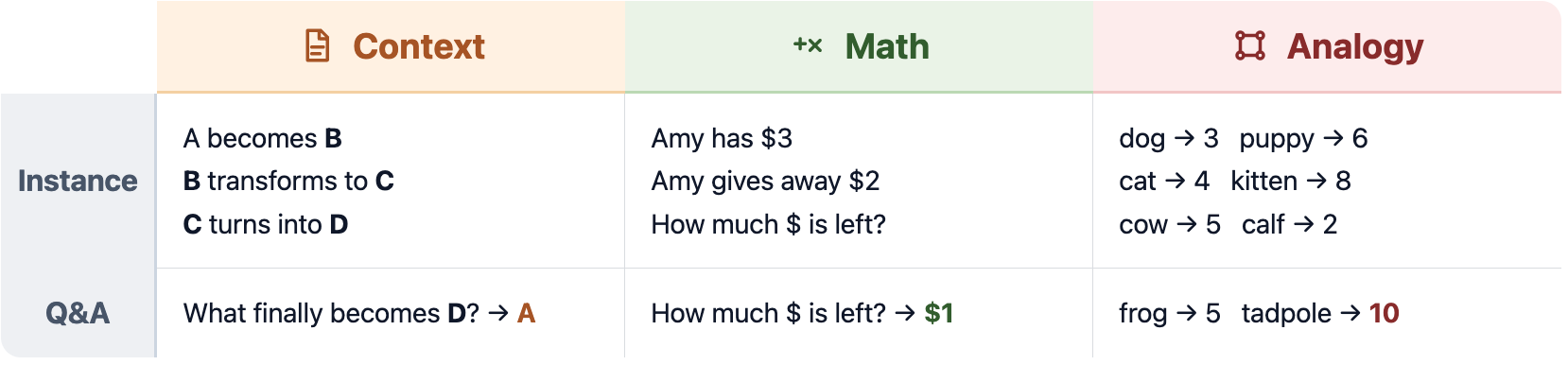}
  \caption{\textbf{Overview of the three task families in CogLoadBench.} We show simplified task formats for clarity; full examples are provided in Appendix~\ref{app:bench_details}.}
  \label{fig:domains}
\end{figure}

Despite this importance, LLM performance has not been systematically evaluated across different, cognitively grounded forms of extraneous load. We address this gap with CogLoadBench, a controlled benchmark that isolates five presentation factors:
\textit{distractor quantity}, \textit{distractor similarity}, \textit{task interference},
\textit{discontinuity}, and \textit{underspecification}
(Figure~\ref{fig:overview}). For each factor, CogLoadBench builds paired prompts that share the same underlying reasoning task and differ only in how the information is presented. We evaluate these factors across three task families: \textit{contextual} reasoning, \textit{mathematical} reasoning, and \textit{analogical} reasoning (Figure~\ref{fig:domains}). Our evaluation of state-of-the-art open-source models on CogLoadBench reveals that extraneous load broadly and consistently harms model performance.

To characterize clear extraneous load effects at a mechanistic level, we introduce the Extraneous Load Score (\textsc{ELS}), a prompt-time metric designed to estimate a model's internal load before generation. \textsc{ELS} reliably increases under extraneous load manipulations, while showing much smaller changes under the intrinsic-difficulty controls we evaluate. We further show that extraneous load effects extend beyond CogLoadBench to realistic long-context QA settings, including QASPER and CUAD. Finally, we introduce \textsc{ELS}-guided latent steering, an inference-time intervention that yields gains in realistic QA by shifting model representations toward lower-load directions without changing the prompt text or model weights.

\paragraph{Contributions.} Our main contributions can be summarized as follows:
\begin{itemize}[leftmargin=18pt, itemsep=2pt, topsep=0pt, parsep=1pt]
    \item We formalize a framework for extraneous cognitive load in LLMs that isolates prompt presentation effects. 
    \item We introduce \textbf{CogLoadBench}, a systematic benchmark designed to evaluate LLMs under five common forms of extraneous load across three task families, and evaluate five open-source models. 
    \item We propose the \textbf{Extraneous Load Score (ELS)}, a prompt-time diagnostic that estimates a model's internal extraneous load. We validate that \textsc{ELS} recovers known load orderings, generalizes across task families and load settings, and remains relatively robust to intrinsic difficulty.

    \item We introduce \textbf{ELS-guided latent steering}, an inference-time intervention that shifts model representations toward lower-load directions, and show that it yields an average gain of \(+0.42\) F1 points on two long-context QA datasets, above standard activation-steering baselines.

\end{itemize}

\section{Prior work}
A core finding in Cognitive Load Theory (CLT) is that changes in task presentation can increase cognitive load (extraneous) and negatively affect performance \citep{Sweller1988CognitiveLD, clt2}. These findings motivate the five presentation factors we study in LLMs; Appendix~\ref{app:human_hypotheses} details their grounding in human cognition and the resulting hypotheses for CogLoadBench.

\paragraph{Presentation effects in LLMs.} A growing body of work shows that LLM performance is affected by the presentation of information. One of the most examined extraneous load factors is distractor quantity: long-context and needle-in-a-haystack evaluations show that performance typically declines as the amount of irrelevant information increases \citep{bai2024longbenchbilingualmultitaskbenchmark, liu2023lostmiddlelanguagemodels,hsieh2024rulerwhatsrealcontext}. Beyond quantity, distractor content matters: models can be misled by information that appears relevant but does not support the correct answer \citep{gao2026firstdrop,lee2026lostnoise}. Other work studies sensitivity to surface form (e.g., different templates or formatting for the same information), ordering effects (e.g., in-context demonstration order), prompt rewording or paraphrases, and multi-instruction settings where secondary goals interfere with the primary task \citep{sclar2024quantifyinglanguagemodelssensitivity, li2025ordermattersrethinkingprompt, cao2024worstpromptperformancelarge, Hwang_2025}. Together, these studies show that presentation factors can significantly affect model performance \citep{hu-levy-2023-prompting,hu2024auxiliary}. However, most existing work studies one presentation factor at a time, often using different task formats, model sets, and evaluation protocols \citep{kaiser2025cogniloadsyntheticnaturallanguage, fu-etal-2025-exclusion, niu-etal-2025-llama}. This makes it difficult to compare different sources of extraneous load directly, or ask whether different load factors share common mechanistic signatures. CogLoadBench addresses this gap with a single controlled benchmark that supports direct comparison across five extraneous load sources under matched content and shared scoring.
\paragraph{Beyond accuracy and coarse proxies.} Most prior work evaluates presentation effects using task accuracy. Some recent work explicitly attempts to map cognitive load concepts onto LLM behavior and uses coarse internal proxies such as entropy or perplexity as indicators of “load” \citep{yang2025sparsebrainsadaptivebrains, shang2025unitedmindsisolatedagents}. These proxies alone do not distinguish intrinsic task difficulty from presentation-induced load. We address this gap with \textsc{ELS}, a prompt-time activation-based diagnostic trained on controlled presentation changes and evaluated separately for robustness to intrinsic difficulty.

\section{Effects of extraneous load}
\label{sec:effects_el}
\paragraph{Benchmark overview.}
CogLoadBench is a controlled benchmark for isolating \emph{extraneous} cognitive load in LLM prompting. We hold the underlying reasoning problem fixed and vary only \emph{how} the same information is presented. Each base problem is generated from a shared template and expanded into a set of matched prompt variants, each changing one presentation factor at a time.

To test whether extraneous load effects generalize across forms of reasoning, CogLoadBench spans three synthetic task families (Figure~\ref{fig:domains}): multi-hop contextual reasoning (\textsc{Context}), multi-step math word problems (\textsc{Math}), and few-shot analogy completion (\textsc{Analogy}).

Every prompt has the same high-level structure. The context contains $K$ reasoning instances from a single task family: one target and $K{-}1$ distractors. Extraneous load conditions modify this context
in different ways. The prompt then repeats the target question and instructs the model to answer only that question; the model is scored only on its target answer.

A \textsc{Context} instance is a  natural-language reasoning chain. The model must answer a question that links an entity in the first step to one in the final step, using the intermediate steps. A \textsc{Math} instance is a math word problem with a single deterministic numeric answer. The model must output that answer. An \textsc{Analogy} instance consists of example pairs of semantically related entities (e.g., \textit{france:paris}, \textit{japan:tokyo}) annotated with symbolic codes that follow a hidden rule (e.g., suffix \textit{\#}). The model must infer the rule and apply it to a new query
pair.

\paragraph{Extraneous load axes.}
We evaluate five presentation axes (Figure~\ref{fig:overview}), organized into two groups. Three axes (\textit{distractor quantity}, \textit{distractor similarity}, and \textit{task interference}) add competing information around the target. The remaining two axes (\textit{discontinuity} and \textit{underspecification}) make the target reasoning path more complicated.

\begin{enumerate}[leftmargin=18pt, itemsep=3pt, topsep=2pt, parsep=1pt]
 \item \textbf{Distractor quantity.} We vary the number of instances $K$ while holding the target's relative position fixed. Distractors are nested, so higher-$K$ prompts reuse all lower-$K$ distractors and add more.

 \item \textbf{Distractor similarity.} At each $K$, we vary how closely distractors resemble the target while requiring different reasoning. In  \textsc{Math}, distractors use the target's topic but a different
 operation; in \textsc{Context}, they use similar entity types but a different relational structure; in \textsc{Analogy}, they use the same semantic relation type but a different symbolic rule.

 \item \textbf{Task interference.} We require the model to complete a secondary task alongside the target question: three-word sequence maintenance, keyword counting, or arithmetic.

 \item \textbf{Discontinuity.} Using the same target and distractor instances, we compare contiguous ordering against two disruptions: shuffling steps within each instance and shuffling steps across instances.

 \item \textbf{Underspecification.} We replace a fully specified element of the target with one that must be inferred from context, while keeping the problem unambiguous and the answer unchanged.

\end{enumerate}

\smallskip
\noindent For each task family, we generate $N{=}200$ base problems per axis. Distractor quantity and similarity are evaluated as $K$-sweeps; the remaining axes use task-natural fixed-$K$ settings compared against matched baselines. Additional details are in Appendix~\ref{app:bench_details}.

\paragraph{Evaluation details.}
We evaluate five open-source instruction-tuned LLMs under greedy decoding: Qwen3-4B, Qwen3-8B, and Qwen3-14B~\citep{yang2025qwen3technicalreport}, Gemma3-12B~\citep{Kamath2025Gemma3T}, and Llama3.1-8B~\citep{grattafiori2024llama}.
 We report accuracy over the same base problems in each condition, averaged across the five models, with standard errors across base problems. We use open-source models because later analyses require access to internal representations.

\subsection{CogLoadBench evaluation results}

Across CogLoadBench, we ask: \textbf{how does changing only the presentation of a reasoning task affect performance?} The results show that extraneous load is not just about prompt length: some of the largest drops occur when the model must recover a disrupted reasoning path.

\begin{figure}[t]
\centering
\includegraphics[width=\linewidth]{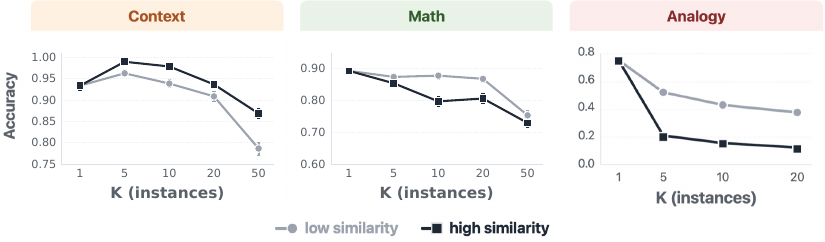}
\caption{\textbf{Distractor quantity and similarity.}
Accuracy averaged across five evaluated models.
Per-model results in Appendix~\ref{app:full_results}.}
\label{fig:quantsim}
\end{figure}

\paragraph{Added content hurts most when it creates competition.}
We first examine load factors that add content around the target: \textit{distractor quantity}, \textit{distractor similarity}, and \textit{task interference}. Increasing \textit{distractor quantity} generally lowers accuracy, but the effect varies by task (Figure~\ref{fig:quantsim}). \textsc{Math} and \textsc{Context} remain relatively robust to moderate increases in \(K\), while \textsc{Analogy} degrades more quickly. This difference reflects what distractors contribute: in \textsc{Math} and \textsc{Context}, they are mostly additional instances to ignore, whereas in \textsc{Analogy}, each distractor provides examples of another possible rule. \textit{Distractor similarity} tests this more directly by holding the number of distractors fixed while making distractors more confusable with the target. High-similarity distractors reduce accuracy in \textsc{Math} and especially \textsc{Analogy}, where they suggest a competing operation or rule, but not in \textsc{Context}, where the question's entities identify the target chain (Figure~\ref{fig:quantsim}). Overall, distractors are most costly when they overlap with the information needed to solve the target.

 \textit{Task interference} shows that competition does not come only from distractor instances. Accuracy generally decreases when the model must complete an additional task alongside the target question (Table~\ref{tab:results-interference}). All three secondary tasks reduce accuracy in \textsc{Analogy}, while counting and arithmetic are more harmful than memory maintenance in \textsc{Math} and \textsc{Context}, suggesting additional interference from active processing in these tasks.

\paragraph{Models struggle when the reasoning path is hard to follow.}
We next consider load factors that make the target reasoning path less clear. \textit{Discontinuity}, which scrambles the order of statements, significantly reduces accuracy even though the task remains solvable (Table~\ref{tab:results-continuity}). This shows that models rely not only on having the right information, but also on seeing it in an order that preserves the reasoning chain. \textit{Underspecification} creates a related problem (Table~\ref{tab:results-underspec}): the answer is still determined by the prompt, but part of the target must be inferred rather than directly stated. This is especially costly in \textsc{Context}, where the model must track entities across a multi-step chain. Together, \textit{discontinuity} and \textit{underspecification} show that extraneous load can arise solely from requiring the model to spend effort understanding how relevant information connects.

\begin{table}[t]
\caption{\textbf{Task interference.}
Averaged across the five models (mean {\scriptsize$\pm$} base-wise SE; $N{=}200$ base problems per model).
Per-model results in Appendix~\ref{app:full_results}.}
\label{tab:results-interference}
\centering
\setlength{\tabcolsep}{4pt}
\begin{tabular}{@{}lcccc@{}}
\\
\toprule
Task & None & Memory & Counting & Arithmetic \\
\midrule
\textsc{Math}    & 88.3{\scriptsize$\pm$1.0} & 88.5{\scriptsize$\pm$1.1} \uab{0.2} & 82.2{\scriptsize$\pm$1.3} \dar{6.1} & 85.1{\scriptsize$\pm$1.1} \dar{3.2} \\
\textsc{Context} & 97.2{\scriptsize$\pm$0.8} & 95.9{\scriptsize$\pm$0.8} \dar{1.3} & 94.8{\scriptsize$\pm$0.8} \dar{2.4} & 94.3{\scriptsize$\pm$0.9} \dar{2.9} \\
\textsc{Analogy} & 45.0{\scriptsize$\pm$2.1} & 38.5{\scriptsize$\pm$1.8} \dar{6.5} & 40.8{\scriptsize$\pm$1.9} \dar{4.2} & 38.6{\scriptsize$\pm$1.9} \dar{6.4} \\
\bottomrule
\end{tabular}
\end{table}
\begin{table}[t]
\caption{\textbf{A harder reasoning path on the same content.}
Accuracy averaged across the five models (mean {\scriptsize$\pm$}
base-wise SE; $N{=}200$ base problems per model). Per-model results in
Appendix~\ref{app:full_results}.}
\label{tab:results-path}
\centering
\small
\setlength{\tabcolsep}{4pt}

\begin{subtable}[t]{0.48\textwidth}\centering
\caption{\textbf{Discontinuity.}}
\label{tab:results-continuity}
\setlength{\tabcolsep}{3pt}
\begin{tabular}{@{}lccc@{}}
\toprule
Task & Ordered & Local shuffle & Global shuffle \\
\midrule
\textsc{Math}    & 89.5{\scriptsize$\pm$0.9} & 76.3{\scriptsize$\pm$1.6} \dar{13.2} & 55.1{\scriptsize$\pm$1.8} \dar{34.4} \\
\textsc{Context} & 97.9{\scriptsize$\pm$0.5} & 95.4{\scriptsize$\pm$0.8} \dar{2.5}  & 83.7{\scriptsize$\pm$1.4} \dar{14.2} \\
\textsc{Analogy} & 49.7{\scriptsize$\pm$2.4} & 29.0{\scriptsize$\pm$1.7} \dar{20.7} & 18.5{\scriptsize$\pm$1.4} \dar{31.2} \\
\bottomrule
\end{tabular}
\end{subtable}
\hfill
\begin{subtable}[t]{0.48\textwidth}\centering
\caption{\textbf{Underspecification.}}
\label{tab:results-underspec}
\begin{tabular}{@{}lcc@{}}
\toprule
Task & Baseline & Underspecified \\
\midrule
\textsc{Math}    & 88.9{\scriptsize$\pm$1.0} & 81.0{\scriptsize$\pm$1.4} \dar{7.9}  \\
\textsc{Context} & 91.3{\scriptsize$\pm$1.0} & 54.4{\scriptsize$\pm$1.9} \dar{36.9} \\
\textsc{Analogy} & 48.6{\scriptsize$\pm$2.2} & 37.3{\scriptsize$\pm$2.1} \dar{11.3}  \\
\bottomrule
\end{tabular}
\end{subtable}

\end{table}

\section{Extraneous Load Score}
\label{els}

\paragraph{Overview.}
Accuracy tells us whether a model answers correctly, but  does not explain \emph{why} a prompt was difficult. LLM failures can come from (1) the difficulty of the underlying reasoning task (intrinsic load), (2) difficulty from how information is presented (extraneous load), and (3) instruction-following failures. Accuracy conflates these factors and thus is limited as a diagnostic metric. 

We therefore introduce the \textbf{Extraneous Load Score (ELS)}, a prompt-time diagnostic for presentation-induced load. \textsc{ELS} is computed from mechanistic indicators in a single forward pass over the prompt. We focus on
prompt-time indicators for two reasons. First, this is motivated by the information-processing stage of problem solving, where load arises while the input is being encoded. Second, it gives a clean, consistent measurement point  before generation dynamics add extra variability.

Because \textsc{ELS} is an internal quantity, there is no ground-truth ``extraneous load'' label we can directly observe. Instead, we evaluate \textsc{ELS} by asking whether it behaves like a measure of extraneous load. First,
when we keep the underlying problem fixed and only change the presentation, \textsc{ELS} should reliably increase, and this pattern should hold across task families and load levels. Second, \textsc{ELS} should be sensitive to extraneous load rather than intrinsic task difficulty, ideally remaining stable when only intrinsic difficulty changes. We follow a construct-validity approach: we evaluate whether \textsc{ELS} responds to the factors it is intended to measure (convergent validity), while remaining comparatively insensitive to intrinsic task difficulty (discriminant validity)
\citep{Cronbach1955ConstructVI, strauss2009construct}.

\paragraph{Set-up.}
We learn \textsc{ELS} from the matched structure of CogLoadBench. Each base problem defines a fixed underlying reasoning task, while low- and high-load variants change only how that task is presented. This lets us train on presentation differences while controlling for intrinsic task difficulty.

\paragraph{Prompt-time feature representation.}
For each prompt, we extract a fixed set of 30 features from internal activations in a single forward pass; no answer generation is required. We split the prompt into a \emph{context} region and a \emph{question} region containing the final query. We compute features over these regions and the full prompt, which includes all prompt tokens. Applying \textsc{ELS} thus only requires identifying the context and query boundaries.

The features capture four properties of prompt encoding: \emph{activation magnitude}, measuring how strongly activations respond; \emph{sparsity and representational spread}, measuring how concentrated or distributed the representation is; \emph{cross-layer stability and drift}, measuring how representations change across layers; and \emph{context--question alignment}, measuring how closely the encoded context aligns with the final query. We average per-layer statistics over early, middle, late, and all layers, and also use the last layer; feature definitions are in Appendix~\ref{app:els_features}. We exclude explicit length features from \textsc{ELS}.

\paragraph{\textsc{ELS} model.}
For each base model \(M\), we learn a scalar scoring function \(f_M : \mathbb{R}^{30} \to \mathbb{R}\) that maps the prompt-time feature vector \(\phi(x)\) to an extraneous load score. We learn a separate function for each model because the same prompt presentation can affect models differently. We implement \(f_M\) as a small two-hidden-layer MLP with dimensions \(64 \to 32 \to 1\) and ReLU activations. For each matched low/high pair \((x_L, x_H)\) from the same base problem, we assign labels \(y(x_L)=0\) and \(y(x_H)=1\), and train with weighted binary cross-entropy:
\[
\mathcal{L}
= \frac{1}{\sum_i w_i} \sum_i w_i \,
\mathrm{BCE}\!\bigl(f_M(\phi(x_i)),\, y(x_i)\bigr),
\]
where \(w_i\) balances load-axis and task-family cells. The resulting score is
\[
\textsc{ELS}_M(x_i) = \sigma(f_M(\phi(x_i))) \in (0,1),
\]
with larger values indicating higher inferred extraneous load.

We use a small classifier so that the signal comes primarily from the prompt-time features rather than model capacity. We test whether \textsc{ELS} behaves like a useful measure of extraneous load.

\begin{table}[t]
\caption{
\textbf{Results for recovering extraneous load on held-out base problems.} Values are ROC AUCs for distinguishing high- from low-load prompts, averaged across five models. Baselines compare against prompt length, mean generated-token entropy, mean top-1 vs.\ top-2 logit margin, and a logistic regression probe on the final query token's hidden states.}
\vspace{0.7em}
\centering
\begin{tabular}{@{}lccccc@{}}
\toprule
Metric & Quantity & Similarity & Interference & Discontinuity & Underspecification \\
\midrule
Prompt length      & \textbf{1.000} & 0.488 & 0.787 & 0.500 & 0.487 \\
Entropy            & 0.492 & 0.526 & 0.623 & 0.606 & 0.545 \\
Logit margin       & 0.537 & 0.523 & 0.583 & 0.603 & 0.544 \\
Hidden-state probe & \textbf{1.000} & 0.777 & \textbf{1.000} & \textbf{1.000} & 0.641 \\
\rowcolor{elsrow}
\textsc{ELS} (ours)         & \textbf{1.000} & \textbf{0.924} & 0.999 & 0.998 & \textbf{0.705} \\
\bottomrule
\end{tabular}
\label{tab:els_baselines_full}
\end{table}

\paragraph{\textsc{ELS} tracks extraneous load better than baselines.}
We first ask whether \textsc{ELS} recovers the intended load ordering on problems unseen during training. For each model, we hold out 20\% of base problems at a time via 5-fold cross-validation, train \(f_M\) only on low/high-load prompts from the training bases, and evaluate on prompts from held-out bases. All variants of a base problem are kept in the same fold across load axes. We report ROC AUC, measuring how well the score distinguishes high- from low-load prompts.

Table~\ref{tab:els_baselines_full} shows that \textsc{ELS} reliably recovers the intended ordering across all five load axes. We compare against prompt length, entropy, and logit margin because prior work uses such quantities as coarse load proxies \citep{yang2025sparsebrainsadaptivebrains, shang2025unitedmindsisolatedagents}. These proxies are limited: prompt length captures distractor quantity only trivially and is much weaker on the other four axes (mean ROC AUC 0.565 vs.\ 0.907 for \textsc{ELS}), while uncertainty metrics like entropy and logit margin do not recover any axis well. We also compare \textsc{ELS} against a stronger representation baseline: a logistic-regression probe trained on PCA-reduced hidden states from the final query token. \textsc{ELS} achieves comparable or higher AUC across the evaluated axes while remaining tied to interpretable properties of prompt encoding.

\paragraph{\textsc{ELS} generalizes across task families and load settings.}
We next test whether \textsc{ELS} transfers across task families. For each model, we train the \textsc{ELS} scorer on two families and evaluate it on the held-out third family, such as training on \textsc{Context} and \textsc{Math} and testing on \textsc{Analogy}. As shown in Table~\ref{tab:els_generalization}(a), performance remains well above chance (70.8--80.5 AUC), suggesting that \textsc{ELS} captures a domain-general notion of extraneous load that extends across qualitatively different reasoning tasks.

We also test whether \textsc{ELS} can generalize to load levels it never sees during training.
For each model and each load factor with more than two levels, we hold out some levels of that factor (e.g., specific values of \(K\), discontinuity levels, or interference types), train on the remaining levels, and evaluate on pairs that include a held-out level. As shown in Table~\ref{tab:els_generalization}(b), \textsc{ELS} continues to recover the correct load ordering on these held-out levels. This shows that once \textsc{ELS} has learned which direction corresponds to “more load” for a factor, it can extend that ordering to new levels of the same factor, behaving like a continuous scale.

\paragraph{\textsc{ELS} quickly calibrates to unseen load factors.}
We also test a stricter leave-one-axis-out setting, where an entire load factor is held out during initial training. We find that \textsc{ELS} generalizes well to distractor quantity and moderately to discontinuity, but performs closer to chance on similarity, interference, and underspecification. This suggests that extraneous load is not captured by a single direction shared across all load factors. However, the same prompt-time features can be adapted to new factors with relatively little data. Adding only four matched low/high-load pairs from a held-out factor increases mean ROC AUC from \(0.650\) to \(0.850\); with 16 pairs, AUC reaches \(0.897\), close to the \(0.930\) achieved when the factor is included during training. Thus, \textsc{ELS} provides a reusable feature representation for diagnosing presentation-induced load, although some factors benefit from light factor-specific calibration (Appendix~\ref{app:els_eval}).

\paragraph{\textsc{ELS} is relatively robust to intrinsic task difficulty.}
Because \textsc{ELS} should measure presentation-induced load rather than task difficulty, we next test whether it changes when only intrinsic difficulty increases. We evaluate paired easy and difficult \textsc{Math} and \textsc{Analogy} problems, keeping the overall prompt structure and answer format fixed. Averaged across five models, accuracy drops by \(16.9\) and \(56.6\) points, respectively. We measure changes in pre-sigmoid \textsc{ELS} scores relative to each model's score gap between matched low- and high-load prompts. The mean signed changes are only \(+6.1\%\) for \textsc{Math} and \(-6.2\%\) for \textsc{Analogy}. Despite substantial accuracy drops, \textsc{ELS} changes little and does not systematically increase with intrinsic difficulty (Appendix~\ref{app:els_intrinsic}).

\paragraph{Different load factors affect different aspects of prompt processing.}
Having established that \textsc{ELS} detects presentation-induced load, we next examine what its features reveal about how each load factor affects prompt processing. We compare the four feature families between matched low- and high-load prompts and ablate each family to measure its contribution to detecting each factor (Appendix~\ref{app:els_family_ablations}). Distractor quantity produces the largest changes: activation magnitude decreases sharply, while sparsity and cross-layer instability increase. Discontinuity produces a similar but weaker pattern, while task interference primarily affects context--question alignment. Similarity and underspecification produce subtler changes: removing stability features lowers similarity AUC by \(11.6\) points, while removing sparsity features lowers underspecification AUC by \(9.8\) points. These differences are consistent with the generalization results: among the non-quantity factors, discontinuity transfers best without calibration, possibly because its effects overlap with those of distractor quantity. These differences also indicate what an intervention should target and how to test whether it moves these features toward their low-load patterns.

\paragraph{Behavior-calibrated \textsc{ELS} compares the impact of presentation changes across factors and models.}
Raw \textsc{ELS} distinguishes higher- from lower-load versions of the same problem, but cannot directly quantify how much different presentation changes hurt accuracy or compare their effects across models. To enable these comparisons, we calibrate \textsc{ELS} against observed accuracy drops: for each task family and model, we fit a linear predictor from prompt-time features to the accuracy drop of each load setting relative to its low-load baseline. On held-out base problems, behavior-calibrated \textsc{ELS} correctly ranks pairs of load settings by their impact on accuracy \(86.8\%\) of the time, compared with \(70.6\%\) for raw \textsc{ELS} differences (Appendix~\ref{app:beh_cal}). Expressing predictions as accuracy drops also lets us compare how strongly different models are affected by the same presentation changes. We use behavior-calibrated \textsc{ELS} to guide steering in Section~\ref{sec:realworld}.

\begin{table*}[t]
\caption{
\textbf{ELS generalization.}
Entries are macro ROC AUCs in percentage points (mean over models).
\textbf{Left:} leave-one-task-out generalization across task families (averaged over load factors).
\textbf{Right:} leave-some-levels-out generalization across load levels for each extraneous load factor (averaged over held-out level conditions and task families).
}
\vspace{0.7em}
\centering
\begin{subtable}[t]{0.47\linewidth}
\centering
\caption{Across tasks}
\label{tab:els_domain_avg}
\begin{tabular}{@{}lcc@{}}
\toprule
Task & All tasks & Held-out task \\
\midrule
\textsc{Context} & 90.6 & 80.5 \dar{10.1} \\
\textsc{Math}    & 89.2 & 76.1 \dar{13.1} \\
\textsc{Analogy} & 95.4 & 70.8 \dar{24.6} \\
\bottomrule
\end{tabular}
\end{subtable}
\hfill
\begin{subtable}[t]{0.47\linewidth}
\centering
\caption{Across load settings}
\label{tab:els_factor_generalization}
\setlength{\tabcolsep}{4pt}
\begin{tabular}{@{}lcc@{}}
\toprule
Load factor   & All settings & Held-out settings \\
\midrule
Quantity      & 100.0 & 100.0 \\
Discontinuity    & 99.9  & 93.2 \dar{6.7} \\
Interference  & 100.0 & 100.0 \\
\bottomrule
\end{tabular}
\end{subtable}
\label{tab:els_generalization}
\end{table*}

\section{Transfer to real-world long-context QA}
\label{sec:realworld}

\begin{table*}[t]
\caption{
\textbf{Real-world extraneous load effects and steering results.}
(a) Mean F1{\scriptsize$\pm$}SE ($N{=}200$), averaged over GPT-5, Claude Sonnet 4.5, Qwen3-14B, and Qwen3-32B. Low vs.\ high load is \(K{=}1\) vs.\ \(K{=}50\) (CUAD) or \(K{=}7\) (QASPER) for quantity, low- vs.\ high-similarity distractors at that \(K\), and no secondary task vs.\ multi-question at \(K{=}5\). Per-model results in Appendix~\ref{app:model-realworld}.
(b) Mean paired F1 change{\scriptsize$\pm$SE} relative to each method's unsteered run, 300 held-out test prompts per setting.
}
\label{tab:realworld_behavior_and_steering}
\centering
\small
\begin{subtable}[t]{\textwidth}
\centering
\caption{\textbf{Real-world effects of extraneous load on QASPER and CUAD.}}
\label{tab:results-realworld-load}
\setlength{\tabcolsep}{4pt}
\begin{tabular}{@{}lcccccc@{}}
\toprule
 & \multicolumn{2}{c}{Quantity} & \multicolumn{2}{c}{Similarity} & \multicolumn{2}{c}{Interference} \\
\cmidrule(lr){2-3}\cmidrule(lr){4-5}\cmidrule(l){6-7}
Dataset & Low load & High load & Low load & High load & Low load & High load \\
\midrule
\textsc{CUAD} & 74.3{\scriptsize$\pm$1.9} & 47.5{\scriptsize$\pm$2.0} \dar{26.9} & 47.5{\scriptsize$\pm$2.0} & 37.2{\scriptsize$\pm$1.8} \dar{10.3} & 68.4{\scriptsize$\pm$2.0} & 56.3{\scriptsize$\pm$1.7} \dar{12.2} \\
\textsc{QASPER} & 45.7{\scriptsize$\pm$1.6} & 23.7{\scriptsize$\pm$1.2} \dar{22.0} & 23.7{\scriptsize$\pm$1.2} & 20.9{\scriptsize$\pm$1.2} \dar{2.7} & 33.3{\scriptsize$\pm$1.6} & 25.1{\scriptsize$\pm$1.2} \dar{8.2} \\
\bottomrule
\end{tabular}
\end{subtable}

\vspace{0.5em}

\begin{subtable}[t]{\textwidth}
\centering
\caption{\textbf{\textsc{ELS}-guided steering and activation-steering baselines on held-out prompts.}}
\label{tab:realworld-steering}
\setlength{\tabcolsep}{4pt}
\begin{tabular}{@{}lllrrrr@{}}
\toprule
Dataset & Model & Axis & \textsc{ELS}-guided & Random & CAA & RepE-PCA \\
\midrule
\textsc{CUAD} & Qwen3-14B & Quantity & $+0.42${\scriptsize$\pm 0.33$} & $-0.06${\scriptsize$\pm 0.32$} & $0.00${\scriptsize$\pm 0.00$} & $0.00${\scriptsize$\pm 0.33$} \\
 &  & Similarity & $-0.24${\scriptsize$\pm 0.17$} & $-0.33${\scriptsize$\pm 0.22$} & $-0.18${\scriptsize$\pm 0.24$} & $-0.12${\scriptsize$\pm 0.17$} \\
 &  & Interference & $+1.39${\scriptsize$\pm 0.98$} & $-1.52${\scriptsize$\pm 1.11$} & $-0.06${\scriptsize$\pm 0.18$} & $-0.37${\scriptsize$\pm 0.40$} \\
 & Qwen3-32B & Quantity & $+0.87${\scriptsize$\pm 0.67$} & $+0.78${\scriptsize$\pm 1.05$} & $+0.18${\scriptsize$\pm 0.19$} & $+0.28${\scriptsize$\pm 0.19$} \\
 &  & Similarity & $+1.76${\scriptsize$\pm 0.71$} & $+1.33${\scriptsize$\pm 1.12$} & $+0.06${\scriptsize$\pm 0.14$} & $-0.74${\scriptsize$\pm 0.51$} \\
 &  & Interference & $+0.20${\scriptsize$\pm 0.36$} & $-0.09${\scriptsize$\pm 0.28$} & $-0.34${\scriptsize$\pm 0.32$} & $-0.46${\scriptsize$\pm 0.30$} \\
\midrule
\textsc{QASPER} & Qwen3-14B & Quantity & $+0.13${\scriptsize$\pm 0.41$} & $-0.08${\scriptsize$\pm 0.52$} & $+0.29${\scriptsize$\pm 0.29$} & $-0.09${\scriptsize$\pm 0.27$} \\
 &  & Similarity & $+0.14${\scriptsize$\pm 0.17$} & $+0.01${\scriptsize$\pm 0.04$} & $0.00${\scriptsize$\pm 0.00$} & $-0.27${\scriptsize$\pm 0.26$} \\
 &  & Interference & $0.00${\scriptsize$\pm 0.03$} & $-0.29${\scriptsize$\pm 0.29$} & $+0.03${\scriptsize$\pm 0.02$} & $-0.36${\scriptsize$\pm 0.31$} \\
 & Qwen3-32B & Quantity & $+0.10${\scriptsize$\pm 0.27$} & $-0.34${\scriptsize$\pm 0.33$} & $+0.01${\scriptsize$\pm 0.01$} & $-0.03${\scriptsize$\pm 0.03$} \\
 &  & Similarity & $-0.11${\scriptsize$\pm 0.07$} & $+0.02${\scriptsize$\pm 0.02$} & $+0.10${\scriptsize$\pm 0.06$} & $-0.02${\scriptsize$\pm 0.07$} \\
 &  & Interference & $+0.37${\scriptsize$\pm 0.25$} & $+0.24${\scriptsize$\pm 0.17$} & $+0.46${\scriptsize$\pm 0.28$} & $+0.11${\scriptsize$\pm 0.27$} \\
\midrule
\multicolumn{3}{l}{Macro-average} & $+0.42$ & $-0.03$ & $+0.05$ & $-0.17$ \\
\bottomrule
\end{tabular}
\end{subtable}
\end{table*}

\paragraph{Motivation.}
CogLoadBench is synthetic by design: it lets us isolate extraneous load under controlled conditions while holding intrinsic task difficulty fixed. We next ask whether the same load effects appear in realistic long-context QA, where models must answer a target question from documents that may contain irrelevant evidence, semantically similar distractors, or competing task demands. This setting naturally mirrors the load axes studied in CogLoadBench and is especially relevant to retrieval-augmented generation (RAG) \citep{lewis2021retrievalaugmentedgenerationknowledgeintensivenlp}. 

We focus on two long-context QA datasets for evaluation: \textbf{QASPER}, which requires answering questions over scientific papers, and \textbf{CUAD}, which requires extracting clauses from legal contracts \citep{dasigi2021dataset,hendrycks2021cuad}.

\paragraph{Experiment design.}
We evaluate realistic analogues of three CogLoadBench axes: \textit{distractor quantity}, \textit{distractor similarity}, and \textit{task interference}. In all cases, we keep the target question and ground-truth answer fixed while changing only the surrounding context.

In QASPER, each prompt starts from an original paper-question pair. For \textit{distractor quantity}, we add randomly sampled distractor papers from the dataset. For \textit{distractor similarity}, we replace random distractors with topically related but answer-irrelevant papers selected by embedding similarity. For \textit{task interference}, we add realistic secondary questions about distractor papers. In CUAD, each prompt starts from a contract question and its ground-truth answer span. We construct quantity and similarity variants by adding random or semantically related cross-contract chunks, and add interference through secondary tasks about distractor chunks (Appendix~\ref{app:realworld_eval}). We do not evaluate \textit{discontinuity} or \textit{underspecification} on either dataset because they are difficult to manipulate without affecting intrinsic task difficulty.

\paragraph{Extraneous load effects persist in real-world settings.}
We first test whether the extraneous load factors isolated in CogLoadBench also affect model behavior on QASPER and CUAD. Table~\ref{tab:realworld_behavior_and_steering}a reports results averaged over two closed and two open models (per-model results in Appendix~\ref{app:model-realworld}). On both datasets, increasing distractor quantity causes large F1 drops across models. Semantically similar distractors and task interference also harm performance, though the magnitude varies by dataset. These results confirm that the extraneous load effects isolated in CogLoadBench arise in realistic long-context QA settings.

\paragraph{ELS-guided latent steering.}
We next ask whether \textsc{ELS} can be used to reduce these load effects. \textsc{ELS}-guided latent steering finds the direction in activation space along which behavior-calibrated \textsc{ELS} increases, and suppresses it at inference time without changing the prompt or the model weights.

For each dataset, we fit behavior-calibrated \textsc{ELS} on the steering-training examples, using F1 drops as the behavioral target (Section~\ref{els}). At each selected layer \(\ell\), a ridge probe from hidden states to this score gives a unit vector \(\hat{w}_\ell\) along which hidden states predict higher load. During context prefill, we suppress the component of each hidden state \(h_{\ell,t}\) along \(\hat{w}_\ell\),
\[
h_{\ell,t}^{\mathrm{steered}}
=
h_{\ell,t}
+
\alpha
\left\langle h_{\ell,t}, \hat{w}_\ell \right\rangle
\hat{w}_\ell,
\qquad
\alpha < 0,
\]
and rescale the state to its original norm. We intervene at middle-to-late layers (\(16,20,24\) for Qwen3-14B; \(24,32,40\) for Qwen3-32B), held fixed across datasets and load axes (Appendix~\ref{app:steering}).

\paragraph{Steering helps reduce extraneous load effects.}
We evaluate \textsc{ELS}-guided steering on Qwen3-14B and Qwen3-32B using QASPER and CUAD. For each model and dataset, we use 150 base examples to fit \textsc{ELS}, the calibrated target, and the steering direction, 200 to select \(\alpha\) for each load axis, and 300 for testing. We report paired F1 changes across all test prompts and compare against three baselines: a random direction with the same layers, steering rule, and strengths; CAA~\citep{panickssery2024steeringllama2contrastive}; and RepE-PCA~\citep{zou2023representation} (details in Appendix~\ref{app:steering}).

Across the 12 settings, \textsc{ELS}-guided steering improves F1 by \(+0.42\) points on average, compared with \(-0.03\) for a random direction, \(+0.05\) for CAA, and \(-0.17\) for RepE-PCA (Table~\ref{tab:realworld-steering}). The largest gains are on CUAD: \(+1.76\) for similarity with Qwen3-32B and \(+1.39\) for interference with Qwen3-14B. The gains come mainly from prompts the model answers poorly without steering; on prompts it already answers correctly, it loses less than a random direction. These results provide initial evidence that \textsc{ELS}-guided steering can improve average performance.

\section{Conclusion}
\label{sec:conclusion}
In this paper, we introduced \textbf{CogLoadBench}, a controlled benchmark for studying extraneous cognitive load from prompt presentation in LLMs. Across synthetic reasoning tasks and realistic long-context QA settings, we showed that presentation of a task alone can substantially affect model performance. We also introduced the \textbf{Extraneous Load Score}, a prompt-time diagnostic for detecting extraneous load before generation, and showed that it is relatively robust to intrinsic difficulty. Finally, we showed that \textbf{ELS-guided latent steering} improves performance at inference time, outperforming standard activation-steering baselines on average.

Together, these results show that extraneous cognitive load is a distinct factor in LLM performance, separate from intrinsic task difficulty. Prior work has mostly studied presentation factors separately; CogLoadBench allows them to be compared within one framework, even though each affects models in different ways. \textsc{ELS} makes this comparison practical. It shows whether a model is broadly sensitive to inefficient presentation or mainly to one factor, and it can be calibrated to new datasets or load types to test whether a mitigation reduces load, as our steering experiments do.

This work has a few limitations. First, across experiments, we generally isolate one load factor at a time, which is useful for understanding specific effects; however, real prompts often combine multiple forms of load, whose effects may reinforce or offset one another. \textsc{ELS} can still compare prompts that contain several forms of load, but it does not show how much each one contributes. Second, our real-world evaluation focuses on long-context QA; other settings, such as multi-turn interaction and agentic workflows, may introduce additional load types. Third, while our load factors come from human cognitive load research, we test them only in LLMs; a matched human study remains future work. Finally, \textsc{ELS} is a diagnostic rather than a complete model of task difficulty, and some domains or load combinations may require additional calibration or additional features. Future work should study interacting load factors, for example whether factor-specific \textsc{ELS} profiles predict how factors combine, and develop mitigations that respond to each prompt's load profile, such as prompt reorganization, retrieval filtering, adaptive inference-time interventions, and training for extraneous load robustness. More broadly, by helping diagnose and reduce failures caused by inefficient prompt presentation, this work supports the development of more robust and reliable LLM systems.

\newpage
\begin{ack}
We thank Shirley Wu, Wanjia Zhao, Batu El, Aneesh Pappu, and the Zou group for helpful feedback on this work. A.G. is supported by the National Science Foundation Graduate Research Fellowship Program under Grant No. DGE-2146755.
\end{ack}

\bibliographystyle{plainnat}

\newpage
\appendix

\section{Appendix overview}
\label{app:overview}

The appendix is organized as follows:

\begin{itemize}
    \item \textbf{Appendix~\ref{app:bench_details}} describes how CogLoadBench is constructed: base problems, task families, the human-study motivation for each load axis, the load axes, and prompt templates.
    
    \item \textbf{Appendix~\ref{app:full_results}} reports full CogLoadBench results, including per-model results.

    \item \textbf{Appendix~\ref{app:els}} describes how \textsc{ELS} is computed, trained, and evaluated, and reports additional \textsc{ELS} analyses.

    \item \textbf{Appendix~\ref{app:realworld_eval}} describes the real-world long-context QA evaluation: datasets, models, load axes, and per-model results.

    \item \textbf{Appendix~\ref{app:steering}} gives implementation details for \textsc{ELS}-guided steering and the activation-steering baselines.

\end{itemize}

\section{CogLoadBench construction}
\label{app:bench_details}
\renewcommand{\thetable}{B\arabic{table}}
\setcounter{table}{0}
\renewcommand{\thefigure}{B\arabic{figure}}
\setcounter{figure}{0}

This section describes how CogLoadBench is constructed: base problems and matched variants (Appendix~\ref{app:base_instances}), task families and problem construction (Appendix~\ref{app:task_families}), the human-study motivation for each load axis (Appendix~\ref{app:human_hypotheses}), the load axes (Appendix~\ref{app:load_axes}), and prompt templates (Appendix~\ref{app:prompt_templates}).

\subsection{Base problems and matched variants}
\label{app:base_instances}

Each evaluation item is a \emph{prompt variant} of a fixed \emph{base problem}. A base problem fixes the underlying reasoning task: its task family, semantic domain, reasoning structure, target question, and answer. The task family determines the type of reasoning required; the semantic domain controls surface content; and the structure controls the reasoning schema or symbolic rule.

For each base problem, we create matched prompt variants by applying one extraneous load manipulation while preserving the target question and answer. This matched design lets us compare low- and high-load presentations of the same underlying reasoning problem. Each prompt contains one target instance and, when applicable, additional distractor instances.

\FloatBarrier
\subsection{Task families and problem construction}
\label{app:task_families}

CogLoadBench contains three task families. In \textsc{Context}, the model follows a five-step reasoning chain to connect its first entity to its last. In \textsc{Math}, the model solves a word problem with one numeric answer. In \textsc{Analogy}, the model infers a hidden rule from example pairs and applies it to a new query.

\begin{table}[h]
\caption{\textbf{Task families.} Semantic domains and reasoning structures in CogLoadBench.}
\centering
\small
\renewcommand{\arraystretch}{1.3}
\setlength{\tabcolsep}{6pt}
\begin{tabular}{@{}p{2.1cm}p{4.0cm}p{6.2cm}@{}}
\toprule
\textbf{Family} & \textbf{Domains} & \textbf{Structures} \\
\midrule

\textsc{Context} &
Chemistry lab, kitchen cooking, carpentry workshop, linguistics, geology &
Forward active chain; passive chain; instrument-tagged chain. \\

\midrule

\textsc{Math} &
Classroom, zoo, kitchen, sports, garden &
Earn--spend--earn; buy--discount--fee; fill--drain--refill. \\

\midrule

\textsc{Analogy} &
Geography, animals, colors, numbers &
Prefix bang \((s\!\to\!!s)\); suffix hash \((s\!\to\!s\#)\); wrap brackets \((s\!\to\![s])\). \\

\bottomrule
\end{tabular}
\label{tab:bench_taxonomy}
\end{table}

Table~\ref{tab:example_instances_table} shows one representative base problem from each task family.

\begin{table}[h]
\centering
\small
\caption{\textbf{Base problems.} Representative examples from the three task families.}
\label{tab:example_instances_table}
\renewcommand{\arraystretch}{1.2}
\begin{tabular}{@{}p{2.7cm}p{8.2cm}p{1.4cm}@{}}
\toprule
\textbf{Task} & \textbf{Instance} & \textbf{Answer} \\
\midrule

\textsc{Context} \newline
\scriptsize Domain: Geology \newline Structure: Passive chain &
1.\; Under compaction and cementation, the sediment Wesalu is lithified to form Wigaha. \newline
2.\; As geothermal heat increases, Wigaha undergoes recrystallization to become Huna. \newline
3.\; Huna transforms through regional metamorphism into Fozeti. \newline
4.\; Fozeti is partially melted and re-solidifies as Cixeda. \newline
5.\; Cixeda evolves into Sexefe. \newline
\textbf{Q:} What is the first sediment in the chain that ultimately produces Sexefe? &
Wesalu \\

\midrule

\textsc{Math} \newline
\scriptsize Domain: Kitchen \newline Structure: buy--discount--fee &
1.\; Julia owns 276 tarts. \newline
2.\; Julia hands out 4 sets of 11 tarts each. \newline
3.\; Julia gets back 19 tarts. \newline
4.\; 10 tarts go missing from Julia. \newline
\textbf{Q:} How many tarts does Julia have now? &
241 \\

\midrule

\textsc{Analogy} \newline
\scriptsize Domain: Geography \newline Structure: Prefix bang &
nevada : ho \quad las vegas : !ho \newline
texas : hi \quad houston : !hi \newline
hawaii : fa \quad honolulu : !fa \newline
\textbf{Q:} new york : go \quad buffalo : ? &
!go \\

\bottomrule
\end{tabular}
\end{table}

\label{app:base_generation}

\paragraph{Context.}
We sample one of five domains, one of three chain structures (Table~\ref{tab:bench_taxonomy}), and an ordered sequence of entities, and prompt GPT-4.1 to express the sequence as a five-step reasoning chain. The question and answer are determined by the sampled chain. Outputs that are malformed, have the wrong number of steps, or do not support the expected answer are rejected and regenerated (fewer than 3\% of generations). We manually reviewed every resulting problem and found no errors.

\paragraph{Math.}
Problems are generated programmatically. We sample one of five domains and one of three arithmetic structures (Table~\ref{tab:bench_taxonomy}), then sample the name, counted object, numbers, and sentence wording. The answer is computed directly from the sampled operations.

\paragraph{Analogy.}
Problems are generated programmatically. We sample one of four domains and one of three symbolic rules (Table~\ref{tab:bench_taxonomy}), then construct three example pairs in which a two-letter token is transformed by the rule. The query applies the same rule to a new token, and the answer is computed from that rule.

\FloatBarrier
\subsection{Motivation from human studies}
\label{app:human_hypotheses}
The five load factors are motivated by findings in the human cognitive-load, attention, and language-comprehension literature. Below, we summarize the relevant findings and state hypotheses for human performance under the corresponding CogLoadBench manipulations.

\paragraph{Distractor quantity.} \citet{harp1998seductive} found that adding interesting but irrelevant details to instructional passages reduced recall of the main ideas and performance on problem-solving transfer tasks. We hypothesize that adding irrelevant instances would reduce accuracy and increase response time, and that the effect would grow with the number of distractors.

\paragraph{Distractor similarity.} \citet{marsh2009interference} found that semantically related irrelevant speech disrupted category-based recall more than unrelated speech, and that this effect depended on the processing required by the recall task. We hypothesize that distractors from the target's domain would cause more errors than distractors from a different domain, potentially including answers drawn from a distractor.

\paragraph{Task interference.} \citet{tombu2011bottleneck} provide neural evidence for a shared attentional bottleneck that limits both perceptual encoding and response selection. We hypothesize that completing a secondary task alongside the target question would increase response time and may reduce accuracy.

\paragraph{Discontinuity.} \citet{chandler1991format} showed that integrating mutually dependent information can improve learning compared with presenting it in separate sources that learners must mentally combine. We hypothesize that shuffling facts would increase integration demands and response time, with interleaving across instances more disruptive than shuffling within an instance.

\paragraph{Underspecification.} \citet{miyake1994ambiguity} showed that working-memory capacity limits readers' ability to keep alternative meanings of ambiguous words active until later disambiguation. Extending this finding to reference resolution, we hypothesize that implicit references requiring additional inference would increase response time and errors relative to explicit references.

These hypotheses provide a basis for a future human comparison with the model results in Section~\ref{sec:effects_el}, examining where behavioral effects align or differ without assuming shared underlying mechanisms.

\FloatBarrier
\subsection{Extraneous load axes}
\label{app:load_axes}

We vary five load axes, one at a time. We randomize the position of the target in the context and keep its relative position fixed across all compared load conditions. Each condition has 200 base problems.

\label{app:axis_examples}
\paragraph{Distractor quantity.}
We add \(K{-}1\) distractor instances around the target, with \(K \in \{1,5,10,20,50\}\) for \textsc{Math} and \textsc{Context} and \(K \in \{1,5,10,20\}\) for \textsc{Analogy}. Distractors are nested: each higher \(K\) keeps all lower-\(K\) distractors and adds more.

\paragraph{Distractor similarity.}
We fix \(K\) and vary how similar the distractors are to the target (Table~\ref{tab:distractor_examples_consistent}). Low-similarity distractors differ from the target in domain and structure; high-similarity distractors share its domain but use a different structure.

\begin{table}[h]
\caption{\textbf{Distractor similarity.} Example distractor types relative to a \textsc{Math} target.}
\centering
\small
\renewcommand{\arraystretch}{1.2}
\begin{tabular}{@{}p{3cm}p{7.8cm}p{1.4cm}@{}}
\toprule
\textbf{Instance type} & \textbf{Instance} & \textbf{Answer} \\
\midrule

Target \newline
\scriptsize Kitchen; buy--discount--fee &
1.\; Julia starts out with 195 scones. \newline
2.\; Julia shares 5 packs of 7 scones each. \newline
3.\; Someone returns 23 scones to Julia. \newline
4.\; 5 scones go missing from Julia. \newline
\textbf{Q:} How many scones does Julia have now? &
178 \\

\midrule

Low similarity \newline
\scriptsize Different domain, different structure &
1.\; Gavin begins with 160 catches. \newline
2.\; Gavin earns 74 catches. \newline
3.\; Gavin gives away 13 catches. \newline
4.\; Gavin collects another 25 catches. \newline
\textbf{Q:} How many catches does Gavin have now? &
246 \\

\midrule

High similarity \newline
\scriptsize Same domain, different structure &
1.\; Initially, Julia has 179 cookies. \newline
2.\; Julia stocks up with 80 cookies. \newline
3.\; Julia gets rid of 34 cookies. \newline
4.\; Another 33 cookies are returned to Julia. \newline
\textbf{Q:} How many cookies does Julia have now? &
258 \\

\bottomrule
\end{tabular}
\label{tab:distractor_examples_consistent}
\end{table}

\paragraph{Task interference.}
We add a secondary task whose answer the model must also report (Figure~\ref{fig:exp4_interference_blocks}): memory, counting, or arithmetic.

\begin{figure}[h]
\centering
\begin{promptbox}[unbreakable]{Interference Tasks}
\begin{Verbatim}[fontsize=\footnotesize, breaklines=true]
Memory:
Before you begin, silently remember the sequence: TIGER, RED, JULY.
At the end, output them under "memory_sequence" in the JSON.

Counting:
While reading the context and final question, count how many times the word
"the" appears. At the end, output the total under "word_count" in the JSON.

Arithmetic:
In your head, compute 34 x 26. Do not show or explain this intermediate
computation. At the end, output the result under "aux_product" in the JSON.
\end{Verbatim}
\end{promptbox}
\caption{\textbf{Task interference.} Each block is prepended to the base prompt and adds one required JSON field.}
\label{fig:exp4_interference_blocks}
\end{figure}

\paragraph{Discontinuity.}
We vary the order of statements (Table~\ref{tab:discontinuity_examples}): natural order (\emph{ordered}), shuffled within each instance (\emph{local shuffle}), or interleaved across instances (\emph{global shuffle}).
\begin{table}[h]
\centering
\small
\caption{\textbf{Discontinuity.} Examples of each condition.}
\label{tab:discontinuity_examples}
\renewcommand{\arraystretch}{1.2}
\begin{tabular}{@{}p{2.7cm}p{9.0cm}@{}}
\toprule
\textbf{Condition} & \textbf{Example prompt order} \\
\midrule

Ordered &
[1.1] Lutu becomes Nogu. \newline
[1.2] Nogu becomes Yomubu. \newline
[1.3] Yomubu becomes Rodipu. \newline
[1.4] Rodipu becomes Tisape. \newline
[1.5] Tisape becomes Reba.
\newline
[2.1] Wesalu becomes Wigaha. \newline
[2.2] Wigaha becomes Huna. \newline
[2.3] Huna becomes Fozeti. \newline
[2.4] Fozeti becomes Cixeda. \newline
[2.5] Cixeda becomes Sexefe. \\

\midrule

Local shuffle &
[1.3] Yomubu becomes Rodipu. \newline
[1.1] Lutu becomes Nogu.\newline
[1.5] Tisape becomes Reba.\newline
[1.2] Nogu becomes Yomubu.\newline
[1.4] Rodipu becomes Tisape.
\newline
[2.4] Fozeti becomes Cixeda.\newline
[2.1] Wesalu becomes Wigaha.\newline
[2.5] Cixeda becomes Sexefe.\newline
[2.2] Wigaha becomes Huna.\newline
[2.3] Huna becomes Fozeti. \\

\midrule

Global shuffle &
[1.2] Nogu becomes Yomubu.\newline
[2.4] Fozeti becomes Cixeda.\newline
[1.5] Tisape becomes Reba.\newline
[2.1] Wesalu becomes Wigaha.\newline
[1.1] Lutu becomes Nogu.\newline
[2.5] Cixeda becomes Sexefe.\newline
[1.3] Yomubu becomes Rodipu.\newline
[2.2] Wigaha becomes Huna.\newline
[1.4] Rodipu becomes Tisape.\newline
[2.3] Huna becomes Fozeti. \\

\bottomrule
\end{tabular}
\end{table}

\paragraph{Underspecification.}
We replace one explicit target element with a reference that must be inferred from context, keeping the answer unambiguous (Table~\ref{tab:underspec_examples}). In \textsc{Math}, a name becomes ``the first person mentioned''; in \textsc{Context}, an intermediate entity becomes ``the latter''; in \textsc{Analogy}, all but one example output is hidden.

\begin{table}[h]
\centering
\small
\caption{\textbf{Underspecification.} We show the relevant prompt text, with changes italicized.}
\label{tab:underspec_examples}
\renewcommand{\arraystretch}{1.2}
\begin{tabular}{@{}p{2.2cm}p{4.5cm}p{4.7cm}@{}}
\toprule
\textbf{Task} & \textbf{Baseline} & \textbf{Underspecified} \\
\midrule

\textsc{Math} &
Yara receives 49 brownies. Yara loses 32 brownies. Yara collects another 23 brownies. &
Yara receives 49 brownies. Yara loses 32 brownies. \emph{The first person mentioned} collects another 23 brownies. \\

\midrule

\textsc{Context} &
Yomubu becomes Rodipu. Rodipu is converted into Tisape. &
Yomubu becomes Rodipu. \emph{The latter} is converted into Tisape. \\

\midrule

\textsc{Analogy} &
nevada : ho \(\to\) las vegas : !ho \newline
texas : hi \(\to\) houston : !hi &
nevada : ho \(\to\) las vegas : ? \newline
texas : hi \(\to\) houston : !hi \\

\bottomrule
\end{tabular}
\end{table}

\FloatBarrier
\subsection{Prompt templates}
\label{app:prompt_templates}

All evaluations use JSON-only outputs. The target question is repeated at the end of the prompt, and the model is instructed to answer only that repeated question. This keeps scoring aligned across conditions even when distractor questions or secondary tasks appear earlier in the prompt.

\begingroup
\begin{promptbox}{Context Prompt Template}
\begin{Verbatim}[fontsize=\footnotesize, breaklines=true]
You will read a series of short 5-step reasoning chains.
Each chain has 5 lines, labeled like [1.1], [1.2], [1.3], [1.4], [1.5] for
chain 1; [2.1], [2.2], [2.3], [2.4], [2.5] for chain 2; and so on.

IMPORTANT:
• Read all chains, but answer ONLY the question that we repeat at the very end.
• Ignore any other questions that appear earlier in the context.

Return your answer in JSON ONLY (no extra text).

[1.1] <s1 for chain 1>
[1.2] <s2 for chain 1>
[1.3] <s3 for chain 1>
[1.4] <s4 for chain 1>
[1.5] <s5 for chain 1>
...
[K.1] <s1 for chain K>
[K.2] <s2 for chain K>
[K.3] <s3 for chain K>
[K.4] <s4 for chain K>
[K.5] <s5 for chain K>

Question to answer (repeated):
<question_only from the TARGET chain>

Answer format (return JSON only). Do not write anything before or after the JSON.
Begin your response on the next line with '{' and output a SINGLE JSON object:
{"reasoning": "<brief explanation of reasoning behind answer>",
 "answer": "<final answer to the repeated question>"}
\end{Verbatim}
\end{promptbox}
\captionof{figure}{\textbf{Prompt template for the \textsc{Context} task family.}}
\label{fig:prompt_context_template}
\endgroup
\par\medskip

\begingroup
\begin{promptbox}{Math Prompt Template}
\begin{Verbatim}[fontsize=\footnotesize, breaklines=true]
You will read a series of short math problems.
Each problem has 5 lines, labeled like [1.1], [1.2], [1.3], [1.4], [1.5] for
problem 1; [2.1], [2.2], [2.3], [2.4], [2.5] for problem 2; and so on.

IMPORTANT:
• Read all problems, but answer ONLY the question that we repeat at the very end.
• Ignore any other questions that appear earlier in the context.

Return your answer in JSON ONLY (no extra text).

[1.1] <line 1 for problem 1>
[1.2] <line 2 for problem 1>
[1.3] <line 3 for problem 1>
[1.4] <line 4 for problem 1>
[1.5] <line 5 for problem 1>
...
[K.1] <line 1 for problem K>
[K.2] <line 2 for problem K>
[K.3] <line 3 for problem K>
[K.4] <line 4 for problem K>
[K.5] <line 5 for problem K>

Question to answer (repeated):
<question_only from the TARGET problem>

Answer format (return JSON only). Do not write anything before or after the JSON.
Begin your response on the next line with '{' and output a SINGLE JSON object:
{"reasoning": "<brief explanation of reasoning behind answer>",
 "answer": "<final numerical answer as a string>"}
\end{Verbatim}
\end{promptbox}
\captionof{figure}{\textbf{Prompt template for the \textsc{Math} task family.}}
\label{fig:prompt_math_template}
\endgroup
\par\medskip

\begingroup
\begin{promptbox}[unbreakable]{Analogy Prompt Template}
\begin{Verbatim}[fontsize=\footnotesize, breaklines=true]
You will see analogy pairs in the format:
  semantic_left : symbolic_code_left
  semantic_right : symbolic_code_right

Pairs that correspond are marked with the same label like [1.1] ... [1.1].
Use labels only to link the two lines of a pair; ignore the numbers when
inferring the rule.

Your job: infer the symbolic transformation that maps the first code to the
second within each semantic pair, then apply the SAME transformation to a
new pair.

The transformation will be ONE of the following patterns (do NOT invent others):
  • prefix_bang   : s -> !s
  • suffix_hash   : s -> s#
  • wrap_brackets : s -> [s]

Examples:

[1.1] <semantic A> : <code A>
[1.1] <semantic B> : <code B>
[2.1] <semantic C> : <code C>
[2.1] <semantic D> : <code D>
...

Now complete a new analogy using the same pattern:

<new semantic left>  : <new code left>
<new semantic right> : ?

Answer format (return JSON only). Do not write anything before or after the JSON.
Begin your response on the next line with '{' and output a SINGLE JSON object:
{"reasoning": "<brief explanation of reasoning behind answer>",
 "answer": "<missing symbolic code>"}
\end{Verbatim}
\end{promptbox}
\captionof{figure}{\textbf{Prompt template for the \textsc{Analogy} task family.}}
\label{fig:prompt_analogy_template}
\endgroup
\par\medskip

\clearpage

\clearpage
\section{CogLoadBench results}
\label{app:full_results}
\renewcommand{\thetable}{C\arabic{table}}
\setcounter{table}{0}

This section reports CogLoadBench accuracy for all five models (Qwen3-4B, Qwen3-8B, Qwen3-14B, Gemma3-12B, Llama3.1-8B) on the three task families and five load axes (Appendix~\ref{app:load_axes}). Table~\ref{tab:cb_macro_distractors} gives model averages for distractor quantity and similarity; model averages for the other three axes are in Tables~\ref{tab:results-interference} and~\ref{tab:results-path}. Tables~\ref{tab:cb_quantity_permodel}--\ref{tab:cb_underspec_permodel} break each axis down by model.

In each table, the first condition column is the baseline, and \dar{} and \uab{} show the change from it. Unless varied, $K=10$ for \textsc{Math} and \textsc{Context} and $K=5$ for \textsc{Analogy}. Each condition uses 200 base problems per model. Standard errors are over base problems for model averages and binomial for per-model results.

\begin{table}[h]
\caption{\textbf{Distractor quantity and similarity.} Averaged over the five models.}
\vspace{0.7em}
\centering

\begin{subtable}[h]{\textwidth}
\centering
\caption{\textbf{Distractor quantity.}}
\label{tab:cb_macro_quantity}
\begin{tabular}{@{}lccccc@{}}
\toprule
Task & $K=1$ & $K=5$ & $K=10$ & $K=20$ & $K=50$ \\
\midrule
\textsc{Math} & 89.3{\scriptsize$\pm$1.0} & 87.4{\scriptsize$\pm$1.0} \dar{1.9} & 87.8{\scriptsize$\pm$1.1} \dar{1.5} & 86.8{\scriptsize$\pm$1.2} \dar{2.5} & 75.3{\scriptsize$\pm$1.4} \dar{14.0} \\
\textsc{Context} & 93.3{\scriptsize$\pm$1.1} & 96.2{\scriptsize$\pm$0.8} \uab{2.9} & 93.8{\scriptsize$\pm$0.9} \uab{0.5} & 90.8{\scriptsize$\pm$1.2} \dar{2.5} & 78.6{\scriptsize$\pm$1.5} \dar{14.7} \\
\textsc{Analogy} & 75.5{\scriptsize$\pm$1.3} & 51.4{\scriptsize$\pm$2.3} \dar{24.1} & 41.9{\scriptsize$\pm$2.1} \dar{33.6} & 38.3{\scriptsize$\pm$2.1} \dar{37.2} & --- \\
\bottomrule
\end{tabular}
\end{subtable}

\vspace{0.8em}

\begin{subtable}[h]{\textwidth}
\centering
\caption{\textbf{Distractor similarity.}}
\label{tab:cb_macro_similarity}
\begin{tabular}{@{}lcc@{}}
\toprule
Task & Low similarity & High similarity \\
\midrule
\textsc{Math} & 87.8{\scriptsize$\pm$1.1} & 79.7{\scriptsize$\pm$1.5} \dar{8.1} \\
\textsc{Context} & 93.8{\scriptsize$\pm$0.9} & 97.8{\scriptsize$\pm$0.6} \uab{4.0} \\
\textsc{Analogy} & 52.3{\scriptsize$\pm$2.2} & 21.0{\scriptsize$\pm$1.8} \dar{31.3} \\
\bottomrule
\end{tabular}
\end{subtable}
\label{tab:cb_macro_distractors}
\end{table}

\begin{table}[h]
\centering
\caption{\textbf{Distractor quantity by model.} Per-model version of Table~\ref{tab:cb_macro_quantity}.}
\vspace{0.7em}
\label{tab:cb_quantity_permodel}
\setlength{\tabcolsep}{4pt}
\begin{tabular}{@{}llccccc@{}}
\toprule
Task & Model & $K=1$ & $K=5$ & $K=10$ & $K=20$ & $K=50$ \\
\midrule
\multirow{5}{*}{\textsc{Math}} & Qwen3-4B & 96.5{\scriptsize$\pm$1.3} & 94.0{\scriptsize$\pm$1.7} \dar{2.5} & 93.0{\scriptsize$\pm$1.8} \dar{3.5} & 90.5{\scriptsize$\pm$2.1} \dar{6.0} & 80.0{\scriptsize$\pm$2.8} \dar{16.5} \\
 & Qwen3-8B & 94.0{\scriptsize$\pm$1.7} & 96.0{\scriptsize$\pm$1.4} \uab{2.0} & 96.0{\scriptsize$\pm$1.4} \uab{2.0} & 93.0{\scriptsize$\pm$1.8} \dar{1.0} & 83.5{\scriptsize$\pm$2.6} \dar{10.5} \\
 & Qwen3-14B & 92.5{\scriptsize$\pm$1.9} & 93.5{\scriptsize$\pm$1.7} \uab{1.0} & 95.5{\scriptsize$\pm$1.5} \uab{3.0} & 93.0{\scriptsize$\pm$1.8} \uab{0.5} & 91.0{\scriptsize$\pm$2.0} \dar{1.5} \\
 & Gemma3-12B & 89.0{\scriptsize$\pm$2.2} & 89.0{\scriptsize$\pm$2.2} & 89.5{\scriptsize$\pm$2.2} \uab{0.5} & 91.5{\scriptsize$\pm$2.0} \uab{2.5} & 78.0{\scriptsize$\pm$2.9} \dar{11.0} \\
 & Llama3.1-8B & 74.5{\scriptsize$\pm$3.1} & 64.5{\scriptsize$\pm$3.4} \dar{10.0} & 65.0{\scriptsize$\pm$3.4} \dar{9.5} & 66.0{\scriptsize$\pm$3.3} \dar{8.5} & 44.0{\scriptsize$\pm$3.5} \dar{30.5} \\
\midrule
\multirow{5}{*}{\textsc{Context}} & Qwen3-4B & 98.5{\scriptsize$\pm$0.9} & 94.0{\scriptsize$\pm$1.7} \dar{4.5} & 91.5{\scriptsize$\pm$2.0} \dar{7.0} & 92.0{\scriptsize$\pm$1.9} \dar{6.5} & 88.0{\scriptsize$\pm$2.3} \dar{10.5} \\
 & Qwen3-8B & 94.0{\scriptsize$\pm$1.7} & 97.0{\scriptsize$\pm$1.2} \uab{3.0} & 98.0{\scriptsize$\pm$1.0} \uab{4.0} & 92.5{\scriptsize$\pm$1.9} \dar{1.5} & 86.5{\scriptsize$\pm$2.4} \dar{7.5} \\
 & Qwen3-14B & 99.0{\scriptsize$\pm$0.7} & 99.0{\scriptsize$\pm$0.7} & 99.0{\scriptsize$\pm$0.7} & 93.5{\scriptsize$\pm$1.7} \dar{5.5} & 91.5{\scriptsize$\pm$2.0} \dar{7.5} \\
 & Gemma3-12B & 95.0{\scriptsize$\pm$1.5} & 97.5{\scriptsize$\pm$1.1} \uab{2.5} & 93.0{\scriptsize$\pm$1.8} \dar{2.0} & 86.0{\scriptsize$\pm$2.5} \dar{9.0} & 60.5{\scriptsize$\pm$3.5} \dar{34.5} \\
 & Llama3.1-8B & 80.0{\scriptsize$\pm$2.8} & 93.5{\scriptsize$\pm$1.7} \uab{13.5} & 87.5{\scriptsize$\pm$2.3} \uab{7.5} & 90.0{\scriptsize$\pm$2.1} \uab{10.0} & 66.5{\scriptsize$\pm$3.3} \dar{13.5} \\
\midrule
\multirow{5}{*}{\textsc{Analogy}} & Qwen3-4B & 54.5{\scriptsize$\pm$3.5} & 38.0{\scriptsize$\pm$3.4} \dar{16.5} & 33.5{\scriptsize$\pm$3.3} \dar{21.0} & 31.0{\scriptsize$\pm$3.3} \dar{23.5} & --- \\
 & Qwen3-8B & 56.5{\scriptsize$\pm$3.5} & 48.5{\scriptsize$\pm$3.5} \dar{8.0} & 39.5{\scriptsize$\pm$3.5} \dar{17.0} & 36.5{\scriptsize$\pm$3.4} \dar{20.0} & --- \\
 & Qwen3-14B & 98.0{\scriptsize$\pm$1.0} & 74.0{\scriptsize$\pm$3.1} \dar{24.0} & 55.5{\scriptsize$\pm$3.5} \dar{42.5} & 50.0{\scriptsize$\pm$3.5} \dar{48.0} & --- \\
 & Gemma3-12B & 79.0{\scriptsize$\pm$2.9} & 52.0{\scriptsize$\pm$3.5} \dar{27.0} & 51.0{\scriptsize$\pm$3.5} \dar{28.0} & 46.5{\scriptsize$\pm$3.5} \dar{32.5} & --- \\
 & Llama3.1-8B & 89.5{\scriptsize$\pm$2.2} & 44.5{\scriptsize$\pm$3.5} \dar{45.0} & 30.0{\scriptsize$\pm$3.2} \dar{59.5} & 27.5{\scriptsize$\pm$3.2} \dar{62.0} & --- \\
\bottomrule
\end{tabular}
\end{table}

\begin{table}[h]
\centering
\caption{\textbf{Distractor similarity by model.} Per-model version of Table~\ref{tab:cb_macro_similarity}.}
\vspace{0.7em}
\label{tab:cb_similarity_permodel}
\begin{tabular}{@{}llcc@{}}
\toprule
Task & Model & Low similarity & High similarity \\
\midrule
\multirow{5}{*}{\textsc{Math}} & Qwen3-4B & 93.0{\scriptsize$\pm$1.8} & 82.5{\scriptsize$\pm$2.7} \dar{10.5} \\
 & Qwen3-8B & 96.0{\scriptsize$\pm$1.4} & 90.0{\scriptsize$\pm$2.1} \dar{6.0} \\
 & Qwen3-14B & 95.5{\scriptsize$\pm$1.5} & 94.0{\scriptsize$\pm$1.7} \dar{1.5} \\
 & Gemma3-12B & 89.5{\scriptsize$\pm$2.2} & 82.5{\scriptsize$\pm$2.7} \dar{7.0} \\
 & Llama3.1-8B & 65.0{\scriptsize$\pm$3.4} & 49.5{\scriptsize$\pm$3.5} \dar{15.5} \\
\midrule
\multirow{5}{*}{\textsc{Context}} & Qwen3-4B & 91.5{\scriptsize$\pm$2.0} & 97.5{\scriptsize$\pm$1.1} \uab{6.0} \\
 & Qwen3-8B & 98.0{\scriptsize$\pm$1.0} & 99.0{\scriptsize$\pm$0.7} \uab{1.0} \\
 & Qwen3-14B & 99.0{\scriptsize$\pm$0.7} & 98.0{\scriptsize$\pm$1.0} \dar{1.0} \\
 & Gemma3-12B & 93.0{\scriptsize$\pm$1.8} & 97.0{\scriptsize$\pm$1.2} \uab{4.0} \\
 & Llama3.1-8B & 87.5{\scriptsize$\pm$2.3} & 97.5{\scriptsize$\pm$1.1} \uab{10.0} \\
\midrule
\multirow{5}{*}{\textsc{Analogy}} & Qwen3-4B & 39.5{\scriptsize$\pm$3.5} & 13.0{\scriptsize$\pm$2.4} \dar{26.5} \\
 & Qwen3-8B & 48.0{\scriptsize$\pm$3.5} & 12.0{\scriptsize$\pm$2.3} \dar{36.0} \\
 & Qwen3-14B & 76.0{\scriptsize$\pm$3.0} & 26.5{\scriptsize$\pm$3.1} \dar{49.5} \\
 & Gemma3-12B & 54.5{\scriptsize$\pm$3.5} & 27.0{\scriptsize$\pm$3.1} \dar{27.5} \\
 & Llama3.1-8B & 43.5{\scriptsize$\pm$3.5} & 26.5{\scriptsize$\pm$3.1} \dar{17.0} \\
\bottomrule
\end{tabular}
\end{table}

\begin{table}[h]
\centering
\caption{\textbf{Task interference by model.} Per-model version of Table~\ref{tab:results-interference}; secondary tasks are shown in Figure~\ref{fig:exp4_interference_blocks}.}
\vspace{0.7em}
\label{tab:cb_interference_permodel}
\begin{tabular}{@{}llcccc@{}}
\toprule
Task & Model & None & Memory & Counting & Arithmetic \\
\midrule
\multirow{5}{*}{\textsc{Math}} & Qwen3-4B & 95.5{\scriptsize$\pm$1.5} & 92.5{\scriptsize$\pm$1.9} \dar{3.0} & 89.5{\scriptsize$\pm$2.2} \dar{6.0} & 86.5{\scriptsize$\pm$2.4} \dar{9.0} \\
 & Qwen3-8B & 94.5{\scriptsize$\pm$1.6} & 93.5{\scriptsize$\pm$1.7} \dar{1.0} & 91.5{\scriptsize$\pm$2.0} \dar{3.0} & 89.5{\scriptsize$\pm$2.2} \dar{5.0} \\
 & Qwen3-14B & 97.0{\scriptsize$\pm$1.2} & 93.5{\scriptsize$\pm$1.7} \dar{3.5} & 86.0{\scriptsize$\pm$2.5} \dar{11.0} & 88.0{\scriptsize$\pm$2.3} \dar{9.0} \\
 & Gemma3-12B & 96.0{\scriptsize$\pm$1.4} & 90.0{\scriptsize$\pm$2.1} \dar{6.0} & 89.0{\scriptsize$\pm$2.2} \dar{7.0} & 91.5{\scriptsize$\pm$2.0} \dar{4.5} \\
 & Llama3.1-8B & 58.5{\scriptsize$\pm$3.5} & 73.0{\scriptsize$\pm$3.1} \uab{14.5} & 55.0{\scriptsize$\pm$3.5} \dar{3.5} & 70.0{\scriptsize$\pm$3.2} \uab{11.5} \\
\midrule
\multirow{5}{*}{\textsc{Context}} & Qwen3-4B & 95.0{\scriptsize$\pm$1.5} & 92.0{\scriptsize$\pm$1.9} \dar{3.0} & 91.0{\scriptsize$\pm$2.0} \dar{4.0} & 85.5{\scriptsize$\pm$2.5} \dar{9.5} \\
 & Qwen3-8B & 97.5{\scriptsize$\pm$1.1} & 97.5{\scriptsize$\pm$1.1} & 94.5{\scriptsize$\pm$1.6} \dar{3.0} & 94.0{\scriptsize$\pm$1.7} \dar{3.5} \\
 & Qwen3-14B & 99.0{\scriptsize$\pm$0.7} & 98.0{\scriptsize$\pm$1.0} \dar{1.0} & 99.0{\scriptsize$\pm$0.7} & 98.0{\scriptsize$\pm$1.0} \dar{1.0} \\
 & Gemma3-12B & 97.5{\scriptsize$\pm$1.1} & 94.5{\scriptsize$\pm$1.6} \dar{3.0} & 96.0{\scriptsize$\pm$1.4} \dar{1.5} & 97.0{\scriptsize$\pm$1.2} \dar{0.5} \\
 & Llama3.1-8B & 97.0{\scriptsize$\pm$1.2} & 97.5{\scriptsize$\pm$1.1} \uab{0.5} & 93.5{\scriptsize$\pm$1.7} \dar{3.5} & 97.0{\scriptsize$\pm$1.2} \\
\midrule
\multirow{5}{*}{\textsc{Analogy}} & Qwen3-4B & 37.5{\scriptsize$\pm$3.4} & 22.0{\scriptsize$\pm$2.9} \dar{15.5} & 42.5{\scriptsize$\pm$3.5} \uab{5.0} & 27.0{\scriptsize$\pm$3.1} \dar{10.5} \\
 & Qwen3-8B & 36.0{\scriptsize$\pm$3.4} & 37.5{\scriptsize$\pm$3.4} \uab{1.5} & 35.0{\scriptsize$\pm$3.4} \dar{1.0} & 37.0{\scriptsize$\pm$3.4} \uab{1.0} \\
 & Qwen3-14B & 63.0{\scriptsize$\pm$3.4} & 55.0{\scriptsize$\pm$3.5} \dar{8.0} & 52.0{\scriptsize$\pm$3.5} \dar{11.0} & 54.0{\scriptsize$\pm$3.5} \dar{9.0} \\
 & Gemma3-12B & 51.5{\scriptsize$\pm$3.5} & 45.5{\scriptsize$\pm$3.5} \dar{6.0} & 44.5{\scriptsize$\pm$3.5} \dar{7.0} & 45.0{\scriptsize$\pm$3.5} \dar{6.5} \\
 & Llama3.1-8B & 37.0{\scriptsize$\pm$3.4} & 32.5{\scriptsize$\pm$3.3} \dar{4.5} & 30.0{\scriptsize$\pm$3.2} \dar{7.0} & 30.0{\scriptsize$\pm$3.2} \dar{7.0} \\
\bottomrule
\end{tabular}
\end{table}

\begin{table}[h]
\centering
\caption{\textbf{Discontinuity by model.} Per-model version of Table~\ref{tab:results-continuity}; conditions are shown in Table~\ref{tab:discontinuity_examples}.}
\vspace{0.7em}
\label{tab:cb_continuity_permodel}
\begin{tabular}{@{}llccc@{}}
\toprule
Task & Model & Ordered & Local shuffle & Global shuffle \\
\midrule
\multirow{5}{*}{\textsc{Math}} & Qwen3-4B & 94.5{\scriptsize$\pm$1.6} & 80.0{\scriptsize$\pm$2.8} \dar{14.5} & 56.0{\scriptsize$\pm$3.5} \dar{38.5} \\
 & Qwen3-8B & 95.0{\scriptsize$\pm$1.5} & 86.5{\scriptsize$\pm$2.4} \dar{8.5} & 70.0{\scriptsize$\pm$3.2} \dar{25.0} \\
 & Qwen3-14B & 98.0{\scriptsize$\pm$1.0} & 89.5{\scriptsize$\pm$2.2} \dar{8.5} & 65.0{\scriptsize$\pm$3.4} \dar{33.0} \\
 & Gemma3-12B & 97.0{\scriptsize$\pm$1.2} & 81.0{\scriptsize$\pm$2.8} \dar{16.0} & 62.0{\scriptsize$\pm$3.4} \dar{35.0} \\
 & Llama3.1-8B & 63.0{\scriptsize$\pm$3.4} & 44.5{\scriptsize$\pm$3.5} \dar{18.5} & 22.5{\scriptsize$\pm$3.0} \dar{40.5} \\
\midrule
\multirow{5}{*}{\textsc{Context}} & Qwen3-4B & 96.5{\scriptsize$\pm$1.3} & 90.5{\scriptsize$\pm$2.1} \dar{6.0} & 69.0{\scriptsize$\pm$3.3} \dar{27.5} \\
 & Qwen3-8B & 98.5{\scriptsize$\pm$0.9} & 99.0{\scriptsize$\pm$0.7} \uab{0.5} & 86.5{\scriptsize$\pm$2.4} \dar{12.0} \\
 & Qwen3-14B & 98.0{\scriptsize$\pm$1.0} & 98.5{\scriptsize$\pm$0.9} \uab{0.5} & 95.0{\scriptsize$\pm$1.5} \dar{3.0} \\
 & Gemma3-12B & 96.5{\scriptsize$\pm$1.3} & 97.5{\scriptsize$\pm$1.1} \uab{1.0} & 87.0{\scriptsize$\pm$2.4} \dar{9.5} \\
 & Llama3.1-8B & 100.0{\scriptsize$\pm$0.0} & 91.5{\scriptsize$\pm$2.0} \dar{8.5} & 81.0{\scriptsize$\pm$2.8} \dar{19.0} \\
\midrule
\multirow{5}{*}{\textsc{Analogy}} & Qwen3-4B & 38.5{\scriptsize$\pm$3.4} & 12.5{\scriptsize$\pm$2.3} \dar{26.0} & 6.5{\scriptsize$\pm$1.7} \dar{32.0} \\
 & Qwen3-8B & 44.5{\scriptsize$\pm$3.5} & 26.0{\scriptsize$\pm$3.1} \dar{18.5} & 15.0{\scriptsize$\pm$2.5} \dar{29.5} \\
 & Qwen3-14B & 68.5{\scriptsize$\pm$3.3} & 35.5{\scriptsize$\pm$3.4} \dar{33.0} & 15.5{\scriptsize$\pm$2.6} \dar{53.0} \\
 & Gemma3-12B & 48.0{\scriptsize$\pm$3.5} & 38.0{\scriptsize$\pm$3.4} \dar{10.0} & 26.0{\scriptsize$\pm$3.1} \dar{22.0} \\
 & Llama3.1-8B & 49.0{\scriptsize$\pm$3.5} & 33.0{\scriptsize$\pm$3.3} \dar{16.0} & 29.5{\scriptsize$\pm$3.2} \dar{19.5} \\
\bottomrule
\end{tabular}
\end{table}

\begin{table}[h]
\centering
\caption{\textbf{Underspecification by model.} Per-model version of Table~\ref{tab:results-underspec}; examples are shown in Table~\ref{tab:underspec_examples}.}
\vspace{0.7em}
\label{tab:cb_underspec_permodel}
\begin{tabular}{@{}llcc@{}}
\toprule
Task & Model & Baseline & Underspecified \\
\midrule
\multirow{5}{*}{\textsc{Math}} & Qwen3-4B & 94.5{\scriptsize$\pm$1.6} & 88.5{\scriptsize$\pm$2.3} \dar{6.0} \\
 & Qwen3-8B & 97.0{\scriptsize$\pm$1.2} & 92.0{\scriptsize$\pm$1.9} \dar{5.0} \\
 & Qwen3-14B & 98.5{\scriptsize$\pm$0.9} & 91.0{\scriptsize$\pm$2.0} \dar{7.5} \\
 & Gemma3-12B & 93.5{\scriptsize$\pm$1.7} & 79.0{\scriptsize$\pm$2.9} \dar{14.5} \\
 & Llama3.1-8B & 61.0{\scriptsize$\pm$3.4} & 54.5{\scriptsize$\pm$3.5} \dar{6.5} \\
\midrule
\multirow{5}{*}{\textsc{Context}} & Qwen3-4B & 86.0{\scriptsize$\pm$2.5} & 42.0{\scriptsize$\pm$3.5} \dar{44.0} \\
 & Qwen3-8B & 91.5{\scriptsize$\pm$2.0} & 45.5{\scriptsize$\pm$3.5} \dar{46.0} \\
 & Qwen3-14B & 97.5{\scriptsize$\pm$1.1} & 74.0{\scriptsize$\pm$3.1} \dar{23.5} \\
 & Gemma3-12B & 89.5{\scriptsize$\pm$2.2} & 55.5{\scriptsize$\pm$3.5} \dar{34.0} \\
 & Llama3.1-8B & 92.0{\scriptsize$\pm$1.9} & 55.0{\scriptsize$\pm$3.5} \dar{37.0} \\
\midrule
\multirow{5}{*}{\textsc{Analogy}} & Qwen3-4B & 45.5{\scriptsize$\pm$3.5} & 30.0{\scriptsize$\pm$3.2} \dar{15.5} \\
 & Qwen3-8B & 41.5{\scriptsize$\pm$3.5} & 32.5{\scriptsize$\pm$3.3} \dar{9.0} \\
 & Qwen3-14B & 72.5{\scriptsize$\pm$3.2} & 52.5{\scriptsize$\pm$3.5} \dar{20.0} \\
 & Gemma3-12B & 44.5{\scriptsize$\pm$3.5} & 32.0{\scriptsize$\pm$3.3} \dar{12.5} \\
 & Llama3.1-8B & 39.0{\scriptsize$\pm$3.4} & 39.5{\scriptsize$\pm$3.5} \uab{0.5} \\
\bottomrule
\end{tabular}
\end{table}

\clearpage
\section{Extraneous Load Score}
\label{app:els}
\renewcommand{\thetable}{D\arabic{table}}
\setcounter{table}{0}

This section describes how \textsc{ELS} is computed (Appendix~\ref{app:els_features}), trained (Appendix~\ref{app:els_setup}), and evaluated, including generalization to new task families, load levels, and load axes (Appendix~\ref{app:els_eval}). We then report a behavior-calibrated version that compares load across settings (Appendix~\ref{app:beh_cal}), robustness to intrinsic difficulty and prompt format (Appendix~\ref{app:els_intrinsic}), feature analyses (Appendix~\ref{app:els_family_ablations}), and how much training data \textsc{ELS} needs (Appendix~\ref{app:els_training_volume}).

\subsection{Prompt-time features}
\label{app:els_features}

For each prompt $x$, we run a single forward pass of the base model and record hidden activations at every layer $\ell \in \mathcal{L} = \{0,\dots,L-1\}$. Let $h_{\ell,t} \in \mathbb{R}^d$ be the residual-stream activation at layer $\ell$ and token position $t$. Using the query boundary, we partition tokens into three regions: a \emph{context} region $\mathcal{C}$ (facts and distractors), a \emph{question} region $\mathcal{Q}$ (the final query), and a \emph{full} region $\mathcal{F}$ comprising all prompt tokens. For each region $R \in \{\mathcal{C},\mathcal{Q},\mathcal{F}\}$ and layer $\ell$, we collect the token matrix
\[
H^{(R)}_\ell
= \bigl[h_{\ell,t}\bigr]_{t \in \mathcal{T}_R} \in \mathbb{R}^{T_R \times d},
\quad
T_R = |\mathcal{T}_R|.
\]

\paragraph{Layer-position aggregation.}
We summarize per-layer activation statistics over early, middle, and late layer groups, the last layer, and all layers pooled together. Let $V_G$ denote the nonmissing entries in an index group $G$. For any scalar per-layer quantity $f^{(R)}_\ell$, we compute:
\[
\begin{aligned}
f^{(R)}_{\text{early}} &= \tfrac{1}{|V_{\mathcal{L}_e}|}\!\!\sum_{\ell \in V_{\mathcal{L}_e}} f^{(R)}_\ell, &
\mathcal{L}_e &= \{0,\dots,\lfloor L/3 \rfloor - 1\}, \\[2pt]
f^{(R)}_{\text{mid}}   &= \tfrac{1}{|V_{\mathcal{L}_m}|}\!\!\sum_{\ell \in V_{\mathcal{L}_m}} f^{(R)}_\ell, &
\mathcal{L}_m &= \{\lfloor L/3 \rfloor,\dots,\lfloor 2L/3 \rfloor - 1\}, \\[2pt]
f^{(R)}_{\text{late}}  &= \tfrac{1}{|V_{\mathcal{L}_l}|}\!\!\sum_{\ell \in V_{\mathcal{L}_l}} f^{(R)}_\ell, &
\mathcal{L}_l &= \{\lfloor 2L/3 \rfloor,\dots,L-1\}, \\[2pt]
f^{(R)}_{\text{last}}  &= f^{(R)}_{L-1}, \\[2pt]
f^{(R)}_{\text{pooled}} &= \tfrac{1}{|V_{\mathcal{L}_p}|}\sum_{\ell\in V_{\mathcal{L}_p}} f^{(R)}_\ell, &
\mathcal{L}_p &= \{0,\dots,L-1\}.
\end{aligned}
\]

\paragraph{1.\ Activation magnitude (5 features).}
For each region $R$, layer $\ell$, and token $t \in \mathcal{T}_R$, we compute the mean token-wise norm
\[
\mu^{(R)}_\ell
= \tfrac{1}{T_R}\!\!\sum_{t \in \mathcal{T}_R} \bigl\| h_{\ell,t} \bigr\|_2.
\]
Selected features: $\mu^{(\mathcal{C})}_{\text{early}}$, $\mu^{(\mathcal{C})}_{\text{pooled}}$, $\mu^{(\mathcal{C})}_{\text{mid}}$, $\mu^{(\mathcal{Q})}_{\text{late}}$, $\mu^{(\mathcal{F})}_{\text{last}}$.

\paragraph{2.\ Sparsity and representational spread (11 features).}

\emph{Gini coefficient} (sparsity of activation magnitudes). For region $R$ and layer $\ell$, flatten the absolute activations $\mathbf{v}^{(R)}_\ell = | H^{(R)}_\ell |_{\text{flattened}} \in \mathbb{R}^{N}$ with $N = T_R \cdot d$, sort ascendingly $0 \le v_{(1)} \le \dots \le v_{(N)}$, and compute
\[
g^{(R)}_\ell
= \frac{2}{N} \cdot
\frac{\sum_{i=1}^{N} i \, v_{(i)}}{\sum_{i=1}^{N} v_{(i)}}
- \frac{N+1}{N}.
\]
Selected features: $g^{(\mathcal{C})}_{\text{mid}}$, $g^{(\mathcal{C})}_{\text{late}}$, $g^{(\mathcal{Q})}_{\text{early}}$, $g^{(\mathcal{Q})}_{\text{mid}}$, $g^{(\mathcal{Q})}_{\text{pooled}}$, $g^{(\mathcal{Q})}_{\text{last}}$.

\emph{Effective rank} (participation ratio). Mean-center token activations and let $\{s_i\}_{i=1}^k$ be the singular values of the centered matrix ($k = \min(T_R,d)$). Define
\[
r^{(R)}_\ell
= \frac{\bigl(\sum_{i=1}^k s_i\bigr)^2}{\sum_{i=1}^k s_i^2}.
\]
Selected features: $r^{(\mathcal{C})}_{\text{early}}$, $r^{(\mathcal{C})}_{\text{mid}}$, $r^{(\mathcal{C})}_{\text{late}}$, $r^{(\mathcal{C})}_{\text{pooled}}$, $r^{(\mathcal{Q})}_{\text{last}}$.\footnote{For computational efficiency, when $T_R > 512$ we uniformly subsample 512 tokens prior to the SVD.}

\paragraph{3.\ Stability and drift across layers (12 features).}
For each region $R$ and layer $\ell$, form the mean activation $\mathbf{m}^{(R)}_\ell = \frac{1}{T_R}\sum_{t \in \mathcal{T}_R} h_{\ell,t}$. For consecutive layers $(\ell,\ell+1)$ we compute three per-step quantities:
\[
\begin{aligned}
d^{(R)}_\ell &= 1 - \cos\!\bigl(\mathbf{m}^{(R)}_\ell, \mathbf{m}^{(R)}_{\ell+1}\bigr) && \text{(drift)},\\
\theta^{(R)}_\ell
&= \arccos\!\bigl(\cos(\mathbf{m}^{(R)}_\ell, \mathbf{m}^{(R)}_{\ell+1})\bigr) \cdot \tfrac{180}{\pi} && \text{(angle, degrees)},\\
M^{(R)}_\ell
&= \bigl\|\mathbf{m}^{(R)}_{\ell+1} - \mathbf{m}^{(R)}_\ell\bigr\|_2 && \text{(step length)}.
\end{aligned}
\]
Selected angle features: $\theta^{(\mathcal{F})}_{\text{mid}}$, $\theta^{(\mathcal{F})}_{\text{late}}$, $\theta^{(\mathcal{C})}_{\text{early}}$, $\theta^{(\mathcal{Q})}_{\text{mid}}$, $\theta^{(\mathcal{Q})}_{\text{late}}$. Selected drift features: $d^{(\mathcal{F})}_{\text{early}}$, $d^{(\mathcal{F})}_{\text{late}}$, $d^{(\mathcal{Q})}_{\text{late}}$. Selected step-length features: $M^{(\mathcal{Q})}_{\text{early}}$, $M^{(\mathcal{Q})}_{\text{mid}}$, $M^{(\mathcal{C})}_{\text{mid}}$, $M^{(\mathcal{F})}_{\text{last}}$.

\paragraph{4.\ Context--question alignment (2 features).}
For each layer $\ell$, compute the cosine similarity between mean context and question representations,
\[
c^{(\mathcal{C},\mathcal{Q})}_\ell
= \frac{\bigl\langle \mathbf{m}^{(\mathcal{C})}_\ell, \mathbf{m}^{(\mathcal{Q})}_\ell\bigr\rangle}
       {\bigl\|\mathbf{m}^{(\mathcal{C})}_\ell\bigr\|_2 \, \bigl\|\mathbf{m}^{(\mathcal{Q})}_\ell\bigr\|_2}.
\]
Selected features: $c^{(\mathcal{C},\mathcal{Q})}_{\text{mid}}$, $c^{(\mathcal{C},\mathcal{Q})}_{\text{late}}$.

\paragraph{Feature selection.}
The four families above (magnitude, sparsity, stability, alignment), expanded over three regions $\{\mathcal{C},\mathcal{Q},\mathcal{F}\}$ and five layer summaries, define a candidate pool. We first remove features whose absolute Spearman correlation with another feature exceeds 0.98, then rank the rest by the change in cross-validated ROC AUC when each is left out. We remove the 20 least important features, leaving 30; in a sweep over the number of features, held-out ROC AUC improves quickly and then plateaus at about this size. Token-length features are excluded throughout.

\subsection{Model and training}
\label{app:els_setup}

\paragraph{Goal.}
For each base model $M$ we learn a scalar scorer $f_M : \mathbb{R}^{30} \to \mathbb{R}$ on the prompt-time feature vector $\phi(x) \in \mathbb{R}^{30}$ (Appendix~\ref{app:els_features}), with larger $\textsc{ELS}_M(x) = \sigma\!\bigl(f_M(\phi(x))\bigr) \in (0,1)$ indicating higher extraneous load.

\paragraph{Training pairs.}
Each training pair compares two prompts from the same base problem that differ in one load manipulation; the lower-load prompt is $x_{\text{low}}$ and the higher-load prompt is $x_{\text{high}}$.
\begin{itemize}
  \item \textbf{Distractor quantity:} fewer vs.\ more distractors ($K_1 < K_2$).
  \item \textbf{Distractor similarity:} low vs.\ high similarity, with $K$ fixed.
  \item \textbf{Task interference:} no secondary task vs.\ a secondary task.
  \item \textbf{Discontinuity:} ordered vs.\ global shuffle.
  \item \textbf{Underspecification:} fully specified vs.\ underspecified.
\end{itemize}

\paragraph{Features and preprocessing.}
For each variant $x$ we extract the 30-dimensional prompt-time feature vector $\phi(x)$ described in Appendix~\ref{app:els_features}. Missing entries are imputed with the per-feature training-set median, then features are z-scored using training-set means and standard deviations.

\paragraph{Model.}
For each base model $M$ we fit a small two-hidden-layer MLP with ReLU activations and hidden widths $(64, 32)$, followed by a single linear output:
\[
f_M(\phi)
= w^\top \,\mathrm{ReLU}\!\bigl(W_2 \,\mathrm{ReLU}(W_1 \phi + b_1) + b_2\bigr) + c.
\]

\paragraph{Training objective.}
Training labels are binary: $y(x_{\text{low}})\!=\!0$, $y(x_{\text{high}})\!=\!1$. We minimize a sample-weighted binary cross-entropy on the logits,
\[
\mathcal{L}
= \frac{1}{\sum_i w_i} \sum_i w_i \cdot \mathrm{BCE}\bigl(f_M(\phi(x_i)),\, y(x_i)\bigr),
\]
where the weights $w_i$ balance contributions across (load axis $\times$ task family) cells so that no single axis or task dominates the fit.

\paragraph{Optimization.}
We optimize with Adam (learning rate $10^{-3}$, weight decay $10^{-3}$) for 200 full-batch epochs, with a fixed random seed for reproducibility. No early stopping or learning-rate schedule is used.

\subsection{Evaluation}
\label{app:els_eval}

\paragraph{Held-out base problems.}
For each base LLM, we train one \(\textsc{ELS}_M\) using low/high-load prompts from all three task families and all five load axes. We evaluate with 5-fold cross-validation grouped by base problem within each task family, so all variants of the same base problem appear in the same fold across load axes.

The main metric is \emph{ROC AUC}, measuring how well the score distinguishes high- from low-load prompts. We compute this metric per model and load axis, then average across the five base models.

\paragraph{Hidden-state baseline.}
For the hidden-state baseline, we concatenate the final query token's representations across layers, standardize them, reduce them to 30 principal components, and fit a logistic regression classifier for each model and load axis, using 5-fold cross-validation grouped by base problem.

\paragraph{Generalization across task families.}
To test whether \(\textsc{ELS}\) transfers across reasoning families, we use a leave-one-task-out protocol. For each held-out family \(t \in \{\textsc{Math}, \textsc{Context}, \textsc{Analogy}\}\), we train \(\textsc{ELS}_M\) only on pairs from the other two families and evaluate on pairs from \(t\). We report ROC AUC on the held-out family, averaged over load axes and models. We compare this to the all-task cross-validated setting as an in-distribution reference.

\paragraph{Generalization across load settings.}
We also test whether \(\textsc{ELS}\) can rank load settings that were not seen during training. For each evaluated load factor, we hold out some settings, train on the remaining settings, and evaluate only on pairs involving the held-out settings. Base problems remain grouped across train and test splits.

\begin{itemize}
    \item \textbf{Distractor quantity.} We evaluate six configurations formed by holding out two of the four non-baseline levels \(K \in \{5,10,20,50\}\), retaining \(K=1\) in both training and evaluation.

    \item \textbf{Discontinuity.} We use three ordering levels: ordered, local shuffle, and global shuffle. We hold out each level in turn and evaluate on pairs of adjacent levels that include the held-out level.

    \item \textbf{Task interference.} We hold out each non-baseline interference type in turn. For each split, we train on the remaining interference types and evaluate on pairs comparing no interference against the held-out type.
\end{itemize}

We do not run this level-holdout analysis for distractor similarity or underspecification, since these axes have only two levels.

\paragraph{Generalization to unseen load axes.}
We next test a stricter setting: leaving out an entire load axis during training. For each held-out extraneous load axis \(a\) (quantity, similarity, interference, discontinuity, underspecification), we train \(\textsc{ELS}_M\) on low/high-load prompts from the other four axes and evaluate on prompts from \(a\). We hold out 20\% of base problems for evaluation, fixed across calibration sizes and seeds.

Without any pairs from the held-out axis (first column of Table~\ref{tab:loao_fewshot}), transfer is mixed: \(\textsc{ELS}\) generalizes strongly to quantity and moderately to discontinuity, but is closer to chance on similarity, interference, and underspecification. This suggests that extraneous load is not fully captured by a single universal direction shared across all presentation factors.

However, \textsc{ELS} can be quickly calibrated to a new axis. For each held-out axis, we add 4 to 128 matched pairs from that axis to the training data and evaluate on the same held-out prompts. Adding only four pairs increases mean ROC AUC from \(0.650\) to \(0.850\); with 16 pairs, it reaches \(0.897\), close to the \(0.930\) achieved when the axis is seen in training.

\begin{table}[h]
\centering
\caption{
\textbf{Leave-one-axis-out generalization with few-shot calibration.}
ROC AUC on the held-out axis, averaged over five models and three random seeds. Columns give the number of matched pairs from the held-out axis added to training. ``Seen'' is the result when the axis is included in training.
}
\label{tab:loao_fewshot}
\small
\begin{tabular}{@{}lccccccc|c@{}}
\toprule
Held-out axis & \(0\) & \(4\) & \(8\) & \(16\) & \(32\) & \(64\) & \(128\) & Seen \\
\midrule
Quantity & 0.989 & 1.000 & 1.000 & 1.000 & 1.000 & 1.000 & 1.000 & 1.000 \\
Similarity & 0.579 & 0.755 & 0.800 & 0.870 & 0.899 & 0.918 & 0.926 & 0.933 \\
Interference & 0.458 & 0.971 & 0.992 & 0.997 & 0.999 & 0.999 & 1.000 & 1.000 \\
Discontinuity & 0.754 & 0.984 & 0.991 & 0.995 & 0.997 & 0.998 & 0.999 & 0.999 \\
Underspecification & 0.470 & 0.539 & 0.581 & 0.621 & 0.658 & 0.677 & 0.697 & 0.718 \\
\midrule
\textbf{Mean} & \textbf{0.650} & \textbf{0.850} & \textbf{0.873} & \textbf{0.897} & \textbf{0.911} & \textbf{0.919} & \textbf{0.924} & \textbf{0.930} \\
\bottomrule
\end{tabular}
\end{table}
\subsection{Behavior-calibrated \textsc{ELS}}
\label{app:beh_cal}

\paragraph{Motivation.}
Because \textsc{ELS} is trained on matched pairs, it shows whether a prompt has more load than its matched low-load version, but it does not place different load settings on one scale (e.g., whether \(K{=}20\) low-similarity distractors are more severe than \(K{=}5\) high-similarity distractors). Within a task family, where base problems have similar intrinsic difficulty, we use accuracy drops to calibrate \textsc{ELS} onto such a scale.

For each task family \(t\), model \(m\), and load setting \(c\), we define behavioral severity as the accuracy drop relative to the matched low-load setting \(c_0\):
\[
s_{t,m,c}
=
\mathrm{Acc}_{t,m,c_0}
-
\mathrm{Acc}_{t,m,c}.
\]
We use the load settings of four factors: distractor quantity (each \(K\)), distractor similarity (low or high at fixed \(K\)), task interference (each secondary task), and discontinuity (local or global shuffle). A larger \(s_{t,m,c}\) means the setting lowers accuracy more.

We then fit a behavior-calibrated \textsc{ELS}. Let \(\phi(x)\in\mathbb{R}^{30}\) be the same prompt-time feature vector used by \textsc{ELS}. For each task family and model, we train a linear model
\[
\widehat{s}(x)
=
w_{t,m}^{\top}\phi(x)+b_{t,m}
\]
to predict the behavioral severity of the load setting of \(x\).

\paragraph{Evaluation.}
For each task family \(t \in \{\textsc{Math}, \textsc{Context}, \textsc{Analogy}\}\) and model \(m\), we split matched base problems 50/50 into calibration and held-out sets, keeping all variants of the same base problem in the same split. Every load setting appears in both splits.

On the calibration split, we compute \(s_{t,m,c}\) for each setting and train the linear model. This defines the calibration-set severity ranking of settings. On held-out examples, we average the calibrated \textsc{ELS} scores within each setting and ask whether these held-out scores reproduce the calibration-set ranking. We report pair-rank accuracy: the fraction of setting pairs ordered correctly.

\begin{table}[t]
\centering
\small
\caption{
\textbf{Behavior-calibrated \textsc{ELS} ranks load settings by severity.}
Values are pair-rank accuracies over load settings, computed from held-out examples and averaged over five models.
Values in blue show the absolute improvement over raw \(\Delta\textsc{ELS}\).
}
\label{tab:beh_cal_els}
\setlength{\tabcolsep}{7pt}
\begin{tabular}{@{}lcc@{}}
\toprule
Task family & Raw \(\Delta\textsc{ELS}\) & Behavior-calibrated \(\textsc{ELS}\) \\
\midrule
\textsc{Math}    & 76.2 & 91.6 {\uab{15.5}} \\
\textsc{Context} & 68.5 & 89.2 {\uab{20.7}} \\
\textsc{Analogy} & 67.1 & 79.5 {\uab{12.4}} \\
\midrule
Mean & 70.6 & 86.8 {\uab{16.2}} \\
\bottomrule
\end{tabular}
\end{table}

Table~\ref{tab:beh_cal_els} shows that raw \(\Delta\textsc{ELS}\) already ranks load settings by severity above chance on held-out examples. The behavior-calibrated linear model substantially improves this ranking, reaching \(86.8\%\) average pair-rank accuracy. Thus, while \textsc{ELS} is mainly a matched-pair diagnostic, a simple calibration for each task family and model can place load settings on a shared severity scale within a task family.

\subsection{Robustness}
\paragraph{Intrinsic task difficulty.}
\label{app:els_intrinsic}

We construct easy and difficult versions of \textsc{Math} (200 pairs) and \textsc{Analogy} (100 pairs) problems with the same prompt structure and answer format (Table~\ref{tab:intrinsic_examples}). Difficult \textsc{Math} problems state quantities as products, and difficult \textsc{Analogy} problems replace the visible transformation with a letter shift.

\begin{table}[h]
\centering
\caption{
\textbf{Example easy/difficult pairs from the intrinsic-difficulty controls.}
}
\label{tab:intrinsic_examples}
\small
\begin{tabular}{@{}p{0.10\linewidth}p{0.40\linewidth}p{0.40\linewidth}@{}}
\toprule
& \textbf{Math: running total} & \textbf{Analogy: letter-string rule} \\
\midrule
\textbf{Easy} &
\texttt{\small [1.1] Ravi starts with 242 stickers. \newline
[1.2] Ravi gives away 63 stickers. \newline
[1.3] Ravi receives 78 more stickers. \newline
[1.4] Ravi loses 80 stickers. \newline
[1.5] How many stickers does Ravi have now?}
&
\texttt{\small [1.1] apple : abc \newline
[1.1] fruit : [!abc\#] \newline
[1.2] oak : xyz \newline
[1.2] tree : [!xyz\#] \newline
[2.1] guitar : qwm \newline
[2.1] instrument : ?}
\\
\midrule
\textbf{Difficult} &
\texttt{\small [1.1] Ravi starts with 242 stickers. \newline
[1.2] Ravi gives away 7 boxes of 9 stickers each. \newline
[1.3] Ravi receives 6 sets of 13 stickers each. \newline
[1.4] Ravi loses 8 bags of 10 stickers each. \newline
[1.5] How many stickers does Ravi have now?}
&
\texttt{\small [1.1] apple : abc \newline
[1.1] fruit : bcd \newline
[1.2] oak : xyz \newline
[1.2] tree : yza \newline
[2.1] guitar : qwm \newline
[2.1] instrument : ?}
\\
\bottomrule
\end{tabular}
\end{table}

We use \textsc{ELS} scores before the sigmoid to avoid saturation. For each model, we compute how much the score changes from easy to difficult problems, as a percentage of the average score gap between low- and high-load prompts. Thus, \(+100\%\) would mean intrinsic difficulty moves \textsc{ELS} as much as extraneous load does. Difficulty lowers accuracy by \(16.9\) points on \textsc{Math} and \(56.6\) points on \textsc{Analogy}, yet \textsc{ELS} shifts by only \(+6.1\%\) and \(-6.2\%\) of the extraneous load gap (Table~\ref{tab:intrinsic_per_model}). Thus, \textsc{ELS} is relatively robust to intrinsic difficulty and does not systematically move in the extraneous load direction.

\begin{table}[hh]
\centering
\caption{
\textbf{Per-model intrinsic-difficulty results using \textsc{ELS} scores before the sigmoid.}
\(\Delta\)Acc is easy minus difficult; Shift is the score change from easy to difficult, as a percentage of the extraneous load gap.
}
\label{tab:intrinsic_per_model}
\small
\setlength{\tabcolsep}{5pt}
\begin{tabular}{@{}llcccccc@{}}
\toprule
Task & Model & Acc easy & Acc difficult & \(\Delta\)Acc & Easy logit & Difficult logit & Shift (\% gap) \\
\midrule
\multirow{6}{*}{\textsc{Math}}
 & Qwen3-4B    & 98.5 & 75.0 & 23.5 & \(-18.33\) & \(-17.01\) & \(+16.7\%\) \\
 & Qwen3-8B    & 98.0 & 83.5 & 14.5 & \(-17.79\) & \(-17.03\) & \(+10.4\%\) \\
 & Qwen3-14B   & 99.0 & 91.5 & 7.5  & \(-13.63\) & \(-13.96\) & \(-4.4\%\) \\
 & Gemma3-12B  & 76.5 & 69.5 & 7.0  & \(-10.06\) & \(-9.64\)  & \(+5.7\%\) \\
 & Llama3.1-8B & 53.0 & 21.0 & 32.0 & \(-16.22\) & \(-16.04\) & \(+2.0\%\) \\
\cmidrule(l){2-8}
 & \textbf{Mean} & \textbf{85.0} & \textbf{68.1} & \textbf{16.9} & & & \textbf{\(+6.1\%\)} \\
\midrule
\multirow{6}{*}{\textsc{Analogy}}
 & Qwen3-4B    & 87.0 & 1.0 & 86.0 & \(-23.51\) & \(-24.21\) & \(-8.8\%\) \\
 & Qwen3-8B    & 22.0 & 0.0 & 22.0 & \(-23.64\) & \(-23.93\) & \(-3.9\%\) \\
 & Qwen3-14B   & 91.0 & 2.0 & 89.0 & \(-21.40\) & \(-21.60\) & \(-2.7\%\) \\
 & Gemma3-12B  & 83.0 & 1.0 & 82.0 & \(-13.99\) & \(-14.61\) & \(-8.4\%\) \\
 & Llama3.1-8B & 4.0  & 0.0 & 4.0  & \(-17.35\) & \(-17.99\) & \(-7.2\%\) \\
\cmidrule(l){2-8}
 & \textbf{Mean} & \textbf{57.4} & \textbf{0.8} & \textbf{56.6} & & & \textbf{\(-6.2\%\)} \\
\bottomrule
\end{tabular}
\end{table}

\paragraph{Prompt format.}
\label{app:format_sensitivity}

To test whether \textsc{ELS} depends on the CogLoadBench prompt template, we reformat prompts with delimiters, JSON, or Markdown headings, keeping the context, query, and answer-format text unchanged. For each model, \textsc{ELS} is trained on the original format and applied without recalibration. Each task family and axis has 20 low-load and 20 high-load prompts, and scores are averaged over five seeds.

\begin{table}[htbp]
\centering
\small
\caption{\textbf{\textsc{ELS} under alternative prompt formats.} ROC AUC averaged
equally across five models and three task families.}
\label{tab:format_sensitivity}
\begin{tabular}{@{}lrrrr@{}}
\toprule
Load axis & Original & Delimiters & JSON & Markdown \\
\midrule
Quantity & 1.000 & 1.000 & 0.997 & 1.000 \\
Similarity & 0.878 & 0.840 & 0.779 & 0.830 \\
Interference & 1.000 & 0.993 & 0.952 & 0.991 \\
Discontinuity & 1.000 & 0.999 & 0.998 & 0.999 \\
Underspecification & 0.634 & 0.577 & 0.602 & 0.584 \\
\midrule
Macro-average & 0.902 & 0.882 & 0.866 & 0.881 \\
\bottomrule
\end{tabular}
\end{table}

Macro-average AUC changes from 0.902 in the original format to 0.866--0.882 across the alternative formats (Table~\ref{tab:format_sensitivity}). Quantity, interference, and discontinuity change little, while similarity and underspecification are more sensitive, so substantially different formats may benefit from light calibration.

\subsection{Feature analysis}
\paragraph{Feature-family ablations.}
\label{app:els_family_ablations}

We analyze how each prompt-time feature family contributes to \textsc{ELS}. We run two complementary ablations. In the \emph{keep-only} setting, we train \(\textsc{ELS}_M\) using only one feature family, testing whether that family is sufficient to distinguish high- from low-load prompts. In the \emph{drop-one} setting, we remove one feature family from the full feature pool, testing how much that family contributes beyond the others.

\begin{table}[h]
\centering
\caption{\textbf{Prompt-time \textsc{ELS} feature-family ablations.}
Per-axis ROC AUC (\%) under 5-fold cross-validation grouped by base problem, averaged across the five base models.}
\label{tab:els_family_ablations}

\begin{subtable}[t]{0.95\linewidth}
\centering
\caption{\textbf{Keep-only.} Train \(\textsc{ELS}_M\) using only the indicated feature family.}
\label{tab:els_family_keep_only}
\begin{tabular}{@{}llllll@{}}
\toprule
Feature family & Quantity & Similarity & Interference & Discontinuity & Underspecification \\
\midrule
Magnitude  (5)  & 100.0 & 63.9 & 85.9 & 84.3 & 51.8 \\
Sparsity   (11) & 100.0 & 68.9 & 95.4 & 97.7 & \textbf{60.2} \\
Stability  (12) & 100.0 & \textbf{81.9} & \textbf{99.3} & \textbf{97.8} & 59.1 \\
Alignment  (2)  & 99.8  & 56.8 & 84.9 & 82.6 & 49.1 \\
\bottomrule
\end{tabular}
\end{subtable}

\vspace{0.7em}

\begin{subtable}[t]{0.95\linewidth}
\centering
\caption{\textbf{Drop-one.} Remove one feature family from the full 30-feature \textsc{ELS}. The Full row differs slightly from Table~\ref{tab:els_baselines_full} because the ablations use different cross-validation splits.}
\label{tab:els_family_drop_one}
\begin{tabular}{@{}llllll@{}}
\toprule
Dropped family & Quantity & Similarity & Interference & Discontinuity & Underspecification \\
\midrule
None (Full) & 100.0 & 92.6 & 99.9 & 99.8 & 70.6 \\
\midrule
Magnitude  & 100.0 & 90.3 \dar{2.3} & 99.9 & 99.7 \dar{0.2} & 67.2 \dar{3.4} \\
Sparsity   & 100.0 & 85.7 \dar{6.9} & 99.9 & 99.4 \dar{0.4} & 60.8 \dar{9.8} \\
Stability  & 100.0 & 81.0 \dar{11.6} & 99.5 \dar{0.4} & 99.8 \dar{0.1} & 64.8 \dar{5.8} \\
Alignment  & 100.0 & 91.1 \dar{1.5} & 99.9 & 99.8 & 71.5 \uab{0.9} \\
\bottomrule
\end{tabular}
\end{subtable}

\end{table}

Table~\ref{tab:els_family_ablations} shows that the feature families are partly redundant on easier axes but complementary on harder ones. Quantity, interference, and discontinuity are recovered well by several individual families and remain near ceiling when any one family is removed. The families differ most on similarity and underspecification. For similarity, Stability is the strongest single family and removing it causes the largest drop. For underspecification, Sparsity contributes most beyond the other families. Overall, no single family explains \textsc{ELS}: the full score benefits from combining multiple prompt-encoding signals.

\paragraph{Feature-level signatures.}
\label{app:els_feature_signatures}
To better understand what \textsc{ELS} captures, we examine how each
prompt-time feature family changes under different load axes. For each model, load axis, and feature, we compute the standardized mean
difference between matched high-load and low-load variants. We then
average these signed effects within each feature family and across the five models.

\begin{table}[t]
\centering
\caption{
\textbf{Feature-level signatures of each load axis.}
Entries are mean signed Cohen's \(d\) between matched high-load and
low-load variants, averaged over features in each family and over five models. Positive values indicate that the feature family increases under higher load.
}
\label{tab:family_directions}
\small
\begin{tabular}{@{}lcccc@{}}
\toprule
Axis & Magnitude & Sparsity & Stability & Alignment \\
\midrule
Quantity       & $-9.36$ & $+2.61$ & $+2.37$ & $-0.51$ \\
Similarity     & $-0.11$ & $-0.12$ & $+0.14$ & $+0.04$ \\
Interference   & $-0.32$ & $+0.16$ & $-0.30$ & $+0.50$ \\
Discontinuity  & $-0.31$ & $+0.31$ & $+0.10$ & $+0.25$ \\
Underspecification     & $+0.13$ & $-0.03$ & $-0.06$ & $+0.02$ \\
\bottomrule
\end{tabular}
\end{table}

Table~\ref{tab:family_directions} shows that different load axes affect different aspects of prompt encoding. Distractor quantity produces the largest feature-level shift: higher-load prompts reduce activation magnitude while increasing sparsity and cross-layer instability. Similarity has much smaller average shifts at the family level. Interference increases context--question alignment and reduces cross-layer drift. Discontinuity increases sparsity and stability-related features. Underspecification shows the weakest family-level signature, matching its lower recovery performance in Table~\ref{tab:els_baselines_full}.

\subsection{Training data needed}
\paragraph{CogLoadBench.}
\label{app:els_training_volume}
To test how much training data \textsc{ELS} needs, we vary the number of training prompts per model and fold, keeping the 30 features fixed and sampling prompts approximately evenly across tasks, axes, and load labels. We report ROC AUC averaged equally over five axes, five models, and five seeds (Table~\ref{tab:els_training_volume}).

\begin{table}[h]
\centering
\caption{\textbf{Amount of \textsc{ELS} training data.} ROC AUC (\%) by number of training prompts per model.}
\label{tab:els_training_volume}
\small
\begin{tabular}{@{}lrrrrr@{}}
\toprule
Load axis & 100 & 500 & 1,000 & 2,000 & 4,000 \\
\midrule
Quantity & 100.0 & 100.0 & 100.0 & 100.0 & 100.0 \\
Similarity & 75.2 & 87.5 & 90.0 & 91.2 & 91.9 \\
Interference & 95.9 & 99.7 & 99.9 & 99.9 & 99.9 \\
Discontinuity & 96.8 & 99.6 & 99.8 & 99.8 & 99.9 \\
Underspecification & 55.8 & 63.4 & 66.5 & 68.5 & 69.9 \\
\midrule
Macro-average & 84.7 & 90.0 & 91.2 & 91.9 & 92.3 \\
\bottomrule
\end{tabular}
\end{table}

\textsc{ELS} reaches 91.2\% macro AUC with 1,000 training prompts, compared with 92.3\% with 4,000 and 92.6\% with the full data. Gains diminish beyond roughly 1,000--2,000 prompts.

\paragraph{New dataset.}
\label{app:els_target_volume}
To test how much data is needed to apply \textsc{ELS} to a new dataset, we train a new \textsc{ELS} for CUAD and QASPER using 5 to 150 base examples and their low- and high-load prompts, keeping the 30 features fixed. Evaluation uses 300 held-out base examples per load axis. Table~\ref{tab:els_target_volume} reports matched-pair ranking accuracy, averaged over quantity, similarity, interference, and five seeds.

\begin{table}[h]
\centering
\caption{\textbf{Training data needed for a new dataset.} Ranking accuracy (\%) on matched pairs by number of training base examples. The average weights the four dataset and model settings equally.}
\label{tab:els_target_volume}
\small
\begin{tabular}{@{}llrrrrrr@{}}
\toprule
Dataset & Model & 5 & 10 & 25 & 50 & 100 & 150 \\
\midrule
CUAD & Qwen3-14B & 89.4 & 93.2 & 94.9 & 95.9 & 96.5 & 96.5 \\
CUAD & Qwen3-32B & 89.2 & 90.5 & 95.3 & 96.1 & 96.6 & 97.0 \\
QASPER & Qwen3-14B & 82.8 & 84.7 & 89.5 & 90.2 & 91.3 & 92.6 \\
QASPER & Qwen3-32B & 84.3 & 87.5 & 90.0 & 92.1 & 93.3 & 93.9 \\
\midrule
\multicolumn{2}{l}{Macro-average} & 86.4 & 89.0 & 92.4 & 93.6 & 94.4 & 95.0 \\
\bottomrule
\end{tabular}
\end{table}

With 25--50 base examples from the new dataset, mean ranking accuracy reaches 92.4--93.6\%, so \textsc{ELS} can be learned from a small sample.

\section{Real-world evaluation}
\label{app:realworld_eval}
\renewcommand{\thetable}{E\arabic{table}}
\setcounter{table}{0}

This section gives details for the real-world evaluation in Section~\ref{sec:realworld}: datasets and models (Appendix~\ref{app:rw_data}), how each load axis is applied (Appendix~\ref{app:rw_axes}), and per-model results (Appendix~\ref{app:model-realworld}). We evaluate the three axes with direct real-world analogues (distractor quantity, distractor similarity, and task interference) on CUAD (legal contracts) and QASPER (scientific papers). We do not evaluate discontinuity or underspecification, because changing order or specificity in real documents is hard to do without changing intrinsic difficulty.

\subsection{Datasets and models}
\label{app:rw_data}

\paragraph{CUAD.}
We use CUAD \citep{hendrycks2021cuad}, which pairs commercial contracts with expert-annotated legal review questions. Each contract is split into paragraph-level chunks; we keep answerable questions whose gold span lies in a single chunk of about 100 words. Each prompt contains the target chunk and distractor chunks, and the model extracts the span that answers the question. We score predictions with token F1 against the gold span.

\paragraph{QASPER.}
We use QASPER \citep{dasigi2021dataset}, which contains questions about NLP papers with extractive, abstractive, yes/no, and unanswerable answers. Each prompt contains the target paper and distractor papers, and the model answers the target question. We score predictions with answer F1.

\paragraph{Models.}
We evaluate GPT-5~\citep{singh2026openaigpt5card}, Claude Sonnet 4.5~\citep{anthropic2025claudesonnet45}, Qwen3-14B, and Qwen3-32B~\citep{yang2025qwen3technicalreport}. Qwen models are run locally with greedy decoding.

\subsection{Load axes}
\label{app:rw_axes}
As in CogLoadBench, variants share the target question and answer; only the surrounding context changes.

\paragraph{Distractor quantity.}
We add \(K{-}1\) distractors to the target: random contract chunks for CUAD (\(K \in \{1,5,20,50\}\)) and low-similarity papers for QASPER (\(K \in \{1,3,5,7\}\)). \(K=1\) contains only the target and is the low-load condition.

\paragraph{Distractor similarity.}
We fix \(K\) (\(K \in \{5,20,50\}\) for CUAD, \(K \in \{3,5,7\}\) for QASPER) and vary how similar the distractors are to the target, using \texttt{text-embedding-3-small} \citep{openai_text_embedding_3_small} and cosine similarity. In CUAD, high-similarity distractors are the nearest-neighbor chunks from other contracts, excluding gold chunks of the same CUAD category to prevent answer leakage; low-similarity distractors are sampled from other contracts. In QASPER, we embed each paper's title and abstract; high-similarity distractors are the nearest-neighbor papers, excluding papers whose answer is in the same section type as the target's; low-similarity distractors are sampled from the least similar papers. The low-similarity condition uses the same distractors as distractor quantity.

\paragraph{Task interference.}
We fix \(K=5\) and add a secondary task: \emph{summary} (the model also summarizes the distractor contexts) or \emph{multi-question} (the model also answers questions about the distractors). Only the target answer is scored.

\subsection{Per-model results}
\label{app:model-realworld}

Tables~\ref{tab:cuad_full_results} and~\ref{tab:qasper_full_results} report per-model results. Table~\ref{tab:realworld_behavior_and_steering}a averages these over the four models.

\begin{table}[h]
\centering
\caption{\textbf{CUAD results.} Mean token F1{\scriptsize$\pm$SE} over the same 200 prompts for every model and setting. \dar{} and \uab{} are relative to \(K=1\) for quantity, to the quantity setting with the same \(K\) for similarity, and to no interference for interference.}
\label{tab:cuad_full_results}
\footnotesize
\setlength{\tabcolsep}{3.5pt}
\begin{tabular}{@{}lcccc@{}}
\toprule
Setting & GPT-5 & Sonnet 4.5 & Qwen3-14B & Qwen3-32B \\
\midrule
\multicolumn{5}{l}{\emph{Distractor quantity: random distractors}} \\
\quad \(K=1\)  & 68.5{\scriptsize$\pm$2.3} & 78.0{\scriptsize$\pm$2.0} & 74.8{\scriptsize$\pm$2.2} & 76.1{\scriptsize$\pm$2.1} \\
\quad \(K=5\)  & 61.1{\scriptsize$\pm$2.4} \dar{7.4} & 73.5{\scriptsize$\pm$2.3} \dar{4.5} & 71.0{\scriptsize$\pm$2.4} \dar{3.8} & 72.9{\scriptsize$\pm$2.3} \dar{3.2} \\
\quad \(K=20\) & 49.5{\scriptsize$\pm$2.5} \dar{19.0} & 60.2{\scriptsize$\pm$2.6} \dar{17.8} & 59.9{\scriptsize$\pm$2.8} \dar{14.9} & 64.2{\scriptsize$\pm$2.7} \dar{11.9} \\
\quad \(K=50\) & 40.8{\scriptsize$\pm$2.3} \dar{27.6} & 53.6{\scriptsize$\pm$2.6} \dar{24.4} & 45.7{\scriptsize$\pm$2.8} \dar{29.1} & 49.7{\scriptsize$\pm$2.7} \dar{26.4} \\
\midrule
\multicolumn{5}{l}{\emph{Distractor similarity (high): nearest-neighbor distractors}} \\
\quad \(K=5\)  & 50.1{\scriptsize$\pm$2.4} \dar{11.0} & 64.4{\scriptsize$\pm$2.4} \dar{9.1} & 67.5{\scriptsize$\pm$2.6} \dar{3.5} & 69.9{\scriptsize$\pm$2.4} \dar{3.1} \\
\quad \(K=20\) & 35.3{\scriptsize$\pm$2.3} \dar{14.1} & 50.6{\scriptsize$\pm$2.5} \dar{9.6} & 48.3{\scriptsize$\pm$2.8} \dar{11.6} & 55.3{\scriptsize$\pm$2.6} \dar{8.9} \\
\quad \(K=50\) & 29.1{\scriptsize$\pm$2.2} \dar{11.7} & 40.5{\scriptsize$\pm$2.4} \dar{13.1} & 37.4{\scriptsize$\pm$2.5} \dar{8.3} & 41.7{\scriptsize$\pm$2.5} \dar{8.0} \\
\midrule
\multicolumn{5}{l}{\emph{Interference at \(K=5\)}} \\
\quad No interference             & 59.7{\scriptsize$\pm$2.4} & 71.3{\scriptsize$\pm$2.4} & 71.1{\scriptsize$\pm$2.4} & 71.6{\scriptsize$\pm$2.3} \\
\quad Summary interference        & 57.3{\scriptsize$\pm$2.4} \dar{2.4} & 68.8{\scriptsize$\pm$2.4} \dar{2.6} & 64.5{\scriptsize$\pm$2.7} \dar{6.5} & 70.3{\scriptsize$\pm$2.3} \dar{1.2} \\
\quad Multi-question interference & 54.6{\scriptsize$\pm$2.0} \dar{5.1} & 63.1{\scriptsize$\pm$2.2} \dar{8.3} & 53.2{\scriptsize$\pm$2.1} \dar{17.8} & 54.2{\scriptsize$\pm$2.1} \dar{17.4} \\
\bottomrule
\end{tabular}
\end{table}

\begin{table}[h]
\centering
\caption{\textbf{QASPER results.} Mean answer F1{\scriptsize$\pm$SE} on the shared 200-prompt set. GPT-5 uses the official QASPER F1; other models use token F1 against the combined reference spans. \dar{} and \uab{} are relative to \(K=1\) for quantity, to the quantity setting with the same \(K\) for similarity, and to no interference for interference.}
\label{tab:qasper_full_results}
\footnotesize
\setlength{\tabcolsep}{3.5pt}
\begin{tabular}{@{}lcccc@{}}
\toprule
Setting & GPT-5 & Sonnet 4.5 & Qwen3-14B & Qwen3-32B \\
\midrule
\multicolumn{5}{l}{\emph{Distractor quantity: low-similarity distractors}} \\
\quad \(K=1\) & 50.6{\scriptsize$\pm$2.2} & 42.3{\scriptsize$\pm$1.7} & 44.6{\scriptsize$\pm$1.8} & 45.1{\scriptsize$\pm$1.7} \\
\quad \(K=3\) & 41.6{\scriptsize$\pm$2.3} \dar{9.0} & 34.7{\scriptsize$\pm$1.8} \dar{7.6} & 36.1{\scriptsize$\pm$1.9} \dar{8.5} & 36.1{\scriptsize$\pm$1.9} \dar{9.1} \\
\quad \(K=5\) & 41.8{\scriptsize$\pm$2.4} \dar{8.8} & 31.5{\scriptsize$\pm$1.8} \dar{10.8} & 28.0{\scriptsize$\pm$2.0} \dar{16.6} & 31.2{\scriptsize$\pm$1.8} \dar{13.9} \\
\quad \(K=7\) & 37.9{\scriptsize$\pm$2.4} \dar{12.7} & 27.1{\scriptsize$\pm$1.8} \dar{15.2} & 15.5{\scriptsize$\pm$1.6} \dar{29.1} & 14.1{\scriptsize$\pm$1.4} \dar{31.0} \\
\midrule
\multicolumn{5}{l}{\emph{Distractor similarity (high): nearest-neighbor distractors}} \\
\quad \(K=3\) & 37.7{\scriptsize$\pm$2.2} \dar{3.9} & 32.0{\scriptsize$\pm$1.7} \dar{2.6} & 32.8{\scriptsize$\pm$1.9} \dar{3.3} & 32.3{\scriptsize$\pm$1.9} \dar{3.8} \\
\quad \(K=5\) & 37.2{\scriptsize$\pm$2.3} \dar{4.7} & 28.9{\scriptsize$\pm$1.7} \dar{2.6} & 26.0{\scriptsize$\pm$1.8} \dar{2.0} & 27.0{\scriptsize$\pm$1.7} \dar{4.2} \\
\quad \(K=7\) & 33.7{\scriptsize$\pm$2.2} \dar{4.3} & 23.3{\scriptsize$\pm$1.5} \dar{3.8} & 12.1{\scriptsize$\pm$1.3} \dar{3.3} & 14.6{\scriptsize$\pm$1.4} \uab{0.5} \\
\midrule
\multicolumn{5}{l}{\emph{Interference at \(K=5\)}} \\
\quad No interference             & 42.3{\scriptsize$\pm$2.4} & 29.9{\scriptsize$\pm$1.8} & 29.9{\scriptsize$\pm$2.0} & 31.1{\scriptsize$\pm$1.9} \\
\quad Summary interference        & 39.1{\scriptsize$\pm$2.2} \dar{3.2} & 27.5{\scriptsize$\pm$1.6} \dar{2.4} & 28.4{\scriptsize$\pm$1.8} \dar{1.5} & 29.2{\scriptsize$\pm$1.8} \dar{1.9} \\
\quad Multi-question interference & 29.0{\scriptsize$\pm$1.8} \dar{13.3} & 22.0{\scriptsize$\pm$1.3} \dar{7.9} & 24.5{\scriptsize$\pm$1.5} \dar{5.4} & 24.8{\scriptsize$\pm$1.5} \dar{6.3} \\
\bottomrule
\end{tabular}
\end{table}

\clearpage
\section{\textsc{ELS}-guided steering}
\label{app:steering}

\setcounter{table}{0}
\renewcommand{\thetable}{F\arabic{table}}

\setcounter{figure}{0}
\renewcommand{\thefigure}{F\arabic{figure}}

This section gives implementation details for the \textsc{ELS}-guided steering in Section~\ref{sec:realworld} and the activation-steering baselines.

\paragraph{Settings.}
Quantity uses \(K=50\) on CUAD and \(K=5\) on QASPER; similarity uses high-similarity distractors at the same \(K\); interference uses the multi-question condition. Macro-averages weight the 12 settings (2 datasets \(\times\) 2 models \(\times\) 3 axes) equally.

\paragraph{Steering.}
We steer at three middle-to-late layers (16, 20, 24 for Qwen3-14B; 24, 32, 40 for Qwen3-32B) during context prefill, and select the strength on validation from
\[
\alpha \in \{-0.02,-0.05,-0.10,-0.20,-0.30,-0.50,-1.00,-2.00,-3.00\}.
\]
We use ridge \(\lambda=1.0\) for the direction and \(\lambda_{\mathrm{cal}}=1.0\) for the calibration map. Decoding is greedy.

\paragraph{Calibration.}
We train \textsc{ELS} on the steering-training split and assign each load condition \(c\) a severity
\[
s_c = \mathrm{F1}(c_0) - \mathrm{F1}(c),
\]
where \(c_0\) is its low-load reference. A linear map from \(\phi(x)\) and the raw \textsc{ELS} score to \(s_c\) gives the calibrated target \(y_{\mathrm{cal}}(x)\), which is used to fit the steering direction.

\paragraph{Direction.}
For each steering layer \(\ell\), let \(h_\ell(x)\) be the prompt representation and \(\mu_\ell\) its mean over the steering-training split. We fit
\[
w_\ell
=
\arg\min_w
\sum_x
\left(
y_{\mathrm{cal}}(x)
-
w^\top(h_\ell(x)-\mu_\ell)
\right)^2
+
\lambda \|w\|_2^2
\]
and set \(\hat{w}_\ell=w_\ell/\|w_\ell\|_2\). Before norm rescaling, the steering update multiplies the component along \(\hat{w}_\ell\) by \(1+\alpha\): \(-1<\alpha<0\) attenuates it, \(\alpha=-1\) removes it, and \(\alpha<-1\) reverses it.

\paragraph{Baselines.}
We compare \textsc{ELS}-guided steering with three baselines. Random-direction steering uses the same layers, steering rule, and strengths as \textsc{ELS}, so only the direction differs. CAA \citep{panickssery2024steeringllama2contrastive} steers along the mean difference between high- and low-load activations, and RepE-PCA \citep{zou2023representation} steers along the top principal component of matched high- and low-load activation differences. Table~\ref{tab:realworld-steering} reports the results.


\begin{thebibliography}{36}
\providecommand{\natexlab}[1]{#1}
\providecommand{\url}[1]{\texttt{#1}}
\expandafter\ifx\csname urlstyle\endcsname\relax
  \providecommand{\doi}[1]{doi: #1}\else
  \providecommand{\doi}{doi: \begingroup \urlstyle{rm}\Url}\fi

\bibitem[{Anthropic}(2025)]{anthropic2025claudesonnet45}
{Anthropic}.
\newblock {Claude Sonnet} 4.5 system card, 2025.
\newblock URL \url{https://www.anthropic.com/claude-sonnet-4-5-system-card}.
\newblock Accessed 2026-05-06.

\bibitem[Bai et~al.(2024)Bai, Lv, Zhang, Lyu, Tang, Huang, Du, Liu, Zeng, Hou, Dong, Tang, and Li]{bai2024longbenchbilingualmultitaskbenchmark}
Yushi Bai, Xin Lv, Jiajie Zhang, Hongchang Lyu, Jiankai Tang, Zhidian Huang, Zhengxiao Du, Xiao Liu, Aohan Zeng, Lei Hou, Yuxiao Dong, Jie Tang, and Juanzi Li.
\newblock {LongBench}: A bilingual, multitask benchmark for long context understanding.
\newblock In \emph{Proceedings of the 62nd Annual Meeting of the Association for Computational Linguistics (Volume 1: Long Papers)}, pages 3119--3137. Association for Computational Linguistics, 2024.
\newblock \doi{10.18653/v1/2024.acl-long.172}.
\newblock URL \url{https://aclanthology.org/2024.acl-long.172/}.

\bibitem[Cao et~al.(2024)Cao, Cai, Zhang, Zou, and Lam]{cao2024worstpromptperformancelarge}
Bowen Cao, Deng Cai, Zhisong Zhang, Yuexian Zou, and Wai Lam.
\newblock On the worst prompt performance of large language models.
\newblock In \emph{Advances in Neural Information Processing Systems}, volume~37, pages 69022--69042, 2024.
\newblock URL \url{https://proceedings.neurips.cc/paper_files/paper/2024/hash/7fa5a377b7ffabcce43cd00231bb3f9c-Abstract-Conference.html}.

\bibitem[Chandler and Sweller(1991)]{chandler1991format}
Paul Chandler and John Sweller.
\newblock Cognitive load theory and the format of instruction.
\newblock \emph{Cognition and Instruction}, 8\penalty0 (4):\penalty0 293--332, 1991.
\newblock \doi{10.1207/s1532690xci0804_2}.

\bibitem[Clark and Kimmons(2023)]{clt2}
Chad Clark and Royce Kimmons.
\newblock Cognitive load theory.
\newblock \emph{EdTechnica}, 2023.
\newblock \doi{10.59668/371.12980}.

\bibitem[Cronbach and Meehl(1955)]{Cronbach1955ConstructVI}
Lee~J. Cronbach and Paul~E. Meehl.
\newblock Construct validity in psychological tests.
\newblock \emph{Psychological Bulletin}, 52\penalty0 (4):\penalty0 281--302, 1955.
\newblock \doi{10.1037/h0040957}.

\bibitem[Dasigi et~al.(2021)Dasigi, Lo, Beltagy, Cohan, Smith, and Gardner]{dasigi2021dataset}
Pradeep Dasigi, Kyle Lo, Iz~Beltagy, Arman Cohan, Noah~A. Smith, and Matt Gardner.
\newblock A dataset of information-seeking questions and answers anchored in research papers.
\newblock In \emph{Proceedings of the 2021 Conference of the North American Chapter of the Association for Computational Linguistics: Human Language Technologies}, pages 4599--4610. Association for Computational Linguistics, 2021.
\newblock \doi{10.18653/v1/2021.naacl-main.365}.
\newblock URL \url{https://aclanthology.org/2021.naacl-main.365/}.

\bibitem[Fu et~al.(2025)Fu, Qin, Huang, Chen, Zhou, and Long]{fu-etal-2025-exclusion}
Qihang Fu, Yongbin Qin, Ruizhang Huang, Yanping Chen, Yulin Zhou, and Lintao Long.
\newblock Exclusion of thought: Mitigating cognitive load in large language models for enhanced reasoning in multiple-choice tasks.
\newblock In \emph{Proceedings of the 63rd Annual Meeting of the Association for Computational Linguistics (Volume 1: Long Papers)}, pages 21673--21686. Association for Computational Linguistics, 2025.
\newblock \doi{10.18653/v1/2025.acl-long.1051}.
\newblock URL \url{https://aclanthology.org/2025.acl-long.1051/}.

\bibitem[Gao et~al.(2026)Gao, Chen, and Huang]{gao2026firstdrop}
Muhan Gao, Zih-Ching Chen, and Kuan-Hao Huang.
\newblock The first drop of ink: Nonlinear impact of distracting information in long-context reasoning.
\newblock In \emph{Proceedings of the 43rd International Conference on Machine Learning}, volume 306 of \emph{Proceedings of Machine Learning Research}, pages 33019--33038. PMLR, 2026.
\newblock URL \url{https://proceedings.mlr.press/v306/gao26b.html}.

\bibitem[{Gemma Team} et~al.(2025){Gemma Team}, Kamath, Ferret, et~al.]{Kamath2025Gemma3T}
{Gemma Team}, Aishwarya Kamath, Johan Ferret, et~al.
\newblock {Gemma} 3 technical report, 2025.
\newblock URL \url{https://arxiv.org/abs/2503.19786}.

\bibitem[Grattafiori et~al.(2024)Grattafiori, Dubey, Jauhri, et~al.]{grattafiori2024llama}
Aaron Grattafiori, Abhimanyu Dubey, Abhinav Jauhri, et~al.
\newblock The {Llama} 3 herd of models, 2024.
\newblock URL \url{https://arxiv.org/abs/2407.21783}.

\bibitem[Harp and Mayer(1998)]{harp1998seductive}
Shannon~F. Harp and Richard~E. Mayer.
\newblock How seductive details do their damage: A theory of cognitive interest in science learning.
\newblock \emph{Journal of Educational Psychology}, 90\penalty0 (3):\penalty0 414--434, 1998.
\newblock \doi{10.1037/0022-0663.90.3.414}.

\bibitem[Hendrycks et~al.(2021)Hendrycks, Burns, Chen, and Ball]{hendrycks2021cuad}
Dan Hendrycks, Collin Burns, Anya Chen, and Spencer Ball.
\newblock {CUAD}: An expert-annotated {NLP} dataset for legal contract review.
\newblock In \emph{Proceedings of the Neural Information Processing Systems Track on Datasets and Benchmarks}, volume~1, 2021.
\newblock URL \url{https://datasets-benchmarks-proceedings.neurips.cc/paper/2021/hash/6ea9ab1baa0efb9e19094440c317e21b-Abstract.html}.

\bibitem[Hsieh et~al.(2024)Hsieh, Sun, Kriman, Acharya, Rekesh, Jia, Zhang, and Ginsburg]{hsieh2024rulerwhatsrealcontext}
Cheng-Ping Hsieh, Simeng Sun, Samuel Kriman, Shantanu Acharya, Dima Rekesh, Fei Jia, Yang Zhang, and Boris Ginsburg.
\newblock {RULER}: What's the real context size of your long-context language models?
\newblock In \emph{First Conference on Language Modeling}, 2024.
\newblock URL \url{https://arxiv.org/abs/2404.06654}.

\bibitem[Hu and Frank(2024)]{hu2024auxiliary}
Jennifer Hu and Michael~C. Frank.
\newblock Auxiliary task demands mask the capabilities of smaller language models.
\newblock In \emph{First Conference on Language Modeling}, 2024.
\newblock URL \url{https://openreview.net/forum?id=U5BUzSn4tD}.

\bibitem[Hu and Levy(2023)]{hu-levy-2023-prompting}
Jennifer Hu and Roger Levy.
\newblock Prompting is not a substitute for probability measurements in large language models.
\newblock In \emph{Proceedings of the 2023 Conference on Empirical Methods in Natural Language Processing}, pages 5040--5060. Association for Computational Linguistics, 2023.
\newblock \doi{10.18653/v1/2023.emnlp-main.306}.
\newblock URL \url{https://aclanthology.org/2023.emnlp-main.306/}.

\bibitem[Hwang et~al.(2025)Hwang, Kim, Koo, Kang, Bae, and Jung]{Hwang_2025}
Yerin Hwang, Yongil Kim, Jahyun Koo, Taegwan Kang, Hyunkyung Bae, and Kyomin Jung.
\newblock {LLMs} can be easily confused by instructional distractions.
\newblock In \emph{Proceedings of the 63rd Annual Meeting of the Association for Computational Linguistics (Volume 1: Long Papers)}, pages 19483--19496. Association for Computational Linguistics, 2025.
\newblock \doi{10.18653/v1/2025.acl-long.957}.
\newblock URL \url{https://aclanthology.org/2025.acl-long.957/}.

\bibitem[Kaiser et~al.(2026)Kaiser, Frigessi, Ramezani-Kebrya, and Ricaud]{kaiser2025cogniloadsyntheticnaturallanguage}
Daniel Kaiser, Arnoldo Frigessi, Ali Ramezani-Kebrya, and Benjamin Ricaud.
\newblock {CogniLoad}: A synthetic natural language reasoning benchmark with tunable length, intrinsic difficulty, and distractor density.
\newblock In \emph{International Conference on Learning Representations}, 2026.
\newblock URL \url{https://iclr.cc/virtual/2026/poster/10011938}.

\bibitem[Lee et~al.(2026)Lee, Jo, Seo, Lee, and Seo]{lee2026lostnoise}
Seongyun Lee, Yongrae Jo, Minju Seo, Moontae Lee, and Minjoon Seo.
\newblock Lost in the noise: How reasoning models fail with contextual distractors, 2026.
\newblock URL \url{https://arxiv.org/abs/2601.07226}.

\bibitem[Lewis et~al.(2020)Lewis, Perez, Piktus, Petroni, Karpukhin, Goyal, K{\"u}ttler, Lewis, Yih, Rockt{\"a}schel, Riedel, and Kiela]{lewis2021retrievalaugmentedgenerationknowledgeintensivenlp}
Patrick Lewis, Ethan Perez, Aleksandra Piktus, Fabio Petroni, Vladimir Karpukhin, Naman Goyal, Heinrich K{\"u}ttler, Mike Lewis, {Wen-tau} Yih, Tim Rockt{\"a}schel, Sebastian Riedel, and Douwe Kiela.
\newblock Retrieval-augmented generation for knowledge-intensive {NLP} tasks.
\newblock In \emph{Advances in Neural Information Processing Systems}, volume~33, pages 9459--9474, 2020.
\newblock URL \url{https://papers.nips.cc/paper/2020/hash/6b493230205f780e1bc26945df7481e5-Abstract.html}.

\bibitem[Li et~al.(2025)Li, Wang, Wang, and Shang]{li2025ordermattersrethinkingprompt}
Warren Li, Yiqian Wang, Zihan Wang, and Jingbo Shang.
\newblock Order matters: Rethinking prompt construction in in-context learning, 2025.
\newblock URL \url{https://arxiv.org/abs/2511.09700}.

\bibitem[Liu et~al.(2024)Liu, Lin, Hewitt, Paranjape, Bevilacqua, Petroni, and Liang]{liu2023lostmiddlelanguagemodels}
Nelson~F. Liu, Kevin Lin, John Hewitt, Ashwin Paranjape, Michele Bevilacqua, Fabio Petroni, and Percy Liang.
\newblock Lost in the middle: How language models use long contexts.
\newblock \emph{Transactions of the Association for Computational Linguistics}, 12:\penalty0 157--173, 2024.
\newblock \doi{10.1162/tacl_a_00638}.
\newblock URL \url{https://aclanthology.org/2024.tacl-1.9/}.

\bibitem[Marsh et~al.(2009)Marsh, Hughes, and Jones]{marsh2009interference}
John~E. Marsh, Robert~W. Hughes, and Dylan~M. Jones.
\newblock Interference by process, not content, determines semantic auditory distraction.
\newblock \emph{Cognition}, 110\penalty0 (1):\penalty0 23--38, 2009.
\newblock \doi{10.1016/j.cognition.2008.08.003}.

\bibitem[Miyake et~al.(1994)Miyake, Just, and Carpenter]{miyake1994ambiguity}
Akira Miyake, Marcel~Adam Just, and Patricia~A. Carpenter.
\newblock Working memory constraints on the resolution of lexical ambiguity: Maintaining multiple interpretations in neutral contexts.
\newblock \emph{Journal of Memory and Language}, 33\penalty0 (2):\penalty0 175--202, 1994.
\newblock \doi{10.1006/jmla.1994.1009}.

\bibitem[Niu et~al.(2025)Niu, Yuan, Wang, Saghir, and Abdi]{niu-etal-2025-llama}
Jingcheng Niu, Xingdi Yuan, Tong Wang, Hamidreza Saghir, and Amir~H. Abdi.
\newblock {L}lama see, {L}lama do: A mechanistic perspective on contextual entrainment and distraction in {LLM}s.
\newblock In \emph{Proceedings of the 63rd Annual Meeting of the {A}ssociation for {C}omputational {L}inguistics (Volume 1: Long Papers)}, pages 16218--16239. Association for Computational Linguistics, 2025.
\newblock \doi{10.18653/v1/2025.acl-long.791}.
\newblock URL \url{https://aclanthology.org/2025.acl-long.791/}.

\bibitem[{OpenAI}(2026)]{openai_text_embedding_3_small}
{OpenAI}.
\newblock {text-embedding-3-small} model, 2026.
\newblock URL \url{https://developers.openai.com/api/docs/models/text-embedding-3-small}.
\newblock OpenAI API documentation. Accessed 2026-05-07.

\bibitem[Panickssery et~al.(2024)Panickssery, Gabrieli, Schulz, Tong, Hubinger, and Turner]{panickssery2024steeringllama2contrastive}
Nina Panickssery, Nick Gabrieli, Julian Schulz, Meg Tong, Evan Hubinger, and Alexander~Matt Turner.
\newblock Steering {Llama} 2 via contrastive activation addition.
\newblock In \emph{Proceedings of the 62nd Annual Meeting of the Association for Computational Linguistics (Volume 1: Long Papers)}, pages 15504--15522. Association for Computational Linguistics, 2024.
\newblock \doi{10.18653/v1/2024.acl-long.828}.
\newblock URL \url{https://aclanthology.org/2024.acl-long.828/}.

\bibitem[Sclar et~al.(2024)Sclar, Choi, Tsvetkov, and Suhr]{sclar2024quantifyinglanguagemodelssensitivity}
Melanie Sclar, Yejin Choi, Yulia Tsvetkov, and Alane Suhr.
\newblock Quantifying language models' sensitivity to spurious features in prompt design or: {How I learned} to start worrying about prompt formatting.
\newblock In \emph{International Conference on Learning Representations}, 2024.
\newblock URL \url{https://iclr.cc/virtual/2024/poster/18650}.

\bibitem[Shang et~al.(2025)Shang, Liu, Liang, Zhang, Hu, and Guo]{shang2025unitedmindsisolatedagents}
HaoYang Shang, Xuan Liu, Zi~Liang, Jie Zhang, Haibo Hu, and Song Guo.
\newblock United minds or isolated agents? {Exploring} coordination of {LLMs} under cognitive load theory, 2025.
\newblock URL \url{https://arxiv.org/abs/2506.06843}.

\bibitem[Singh et~al.(2026)Singh, Fry, Perelman, et~al.]{singh2026openaigpt5card}
Aaditya Singh, Adam Fry, Adam Perelman, et~al.
\newblock {OpenAI GPT-5} system card, 2026.
\newblock URL \url{https://arxiv.org/abs/2601.03267}.

\bibitem[Strauss and Smith(2009)]{strauss2009construct}
Milton~E. Strauss and Gregory~T. Smith.
\newblock Construct validity: Advances in theory and methodology.
\newblock \emph{Annual Review of Clinical Psychology}, 5:\penalty0 1--25, 2009.
\newblock \doi{10.1146/annurev.clinpsy.032408.153639}.

\bibitem[Sweller(1988)]{Sweller1988CognitiveLD}
John Sweller.
\newblock Cognitive load during problem solving: Effects on learning.
\newblock \emph{Cognitive Science}, 12\penalty0 (2):\penalty0 257--285, 1988.
\newblock \doi{10.1207/s15516709cog1202_4}.

\bibitem[Tombu et~al.(2011)Tombu, Asplund, Dux, Godwin, Martin, and Marois]{tombu2011bottleneck}
Michael~N. Tombu, Christopher~L. Asplund, Paul~E. Dux, Douglass Godwin, Justin~W. Martin, and Ren{\'e} Marois.
\newblock A unified attentional bottleneck in the human brain.
\newblock \emph{Proceedings of the National Academy of Sciences}, 108\penalty0 (33):\penalty0 13426--13431, 2011.
\newblock \doi{10.1073/pnas.1103583108}.

\bibitem[Yang et~al.(2025)Yang, Li, Yang, et~al.]{yang2025qwen3technicalreport}
An~Yang, Anfeng Li, Baosong Yang, et~al.
\newblock Qwen3 technical report, 2025.
\newblock URL \url{https://arxiv.org/abs/2505.09388}.

\bibitem[Yang et~al.(2026)Yang, Wang, Ma, Yu, Chersoni, and Huang]{yang2025sparsebrainsadaptivebrains}
Yiheng Yang, Yujie Wang, Chi Ma, Lei Yu, Emmanuele Chersoni, and Chu-Ren Huang.
\newblock Sparse brains are also adaptive brains: Cognitive-load-aware dynamic activation for {LLMs}.
\newblock In \emph{Findings of the Association for Computational Linguistics: EACL 2026}, pages 5124--5138. Association for Computational Linguistics, 2026.
\newblock \doi{10.18653/v1/2026.findings-eacl.270}.
\newblock URL \url{https://aclanthology.org/2026.findings-eacl.270/}.

\bibitem[Zou et~al.(2023)Zou, Phan, Chen, et~al.]{zou2023representation}
Andy Zou, Long Phan, Sarah Chen, et~al.
\newblock Representation engineering: A top-down approach to {AI} transparency, 2023.
\newblock URL \url{https://arxiv.org/abs/2310.01405}.

\end{thebibliography}
\end{document}